\documentclass[review]{elsarticle}

\usepackage{hyperref}

\journal{Journal of \LaTeX\ Templates}

\biboptions{authoryear}

\usepackage{siunitx}

\usepackage{booktabs}  
\newcolumntype{P}[1]{>{\centering\arraybackslash}p{#1}}
\usepackage{multirow}
\newcommand{\minitab}[2][c]{\begin{tabular}{#1}#2\end{tabular}}

\usepackage{epstopdf}
\AppendGraphicsExtensions{.tif}

\usepackage[acronym,toc,shortcuts]{glossaries}
\glsdisablehyper

\newacronym{AAT}{AAT}{Automatic Aerial Triangulation}
\newacronym{ALS}{ALS}{Airborne Laser Scanning}
\newacronym{CNN}{CNN}{Convolutional Neural Network}
\newacronym[plural=COGs, firstplural=Centers of Gravity (COGs)]{COG}{COG}{Center of Gravity}
\newacronym{DL}{DL}{Deep Learning}
\newacronym{GSD}{GSD}{Ground Sampling Distance}
\newacronym{GT}{GT}{Ground Truth}
\newacronym{H3D}{H3D}{Hessigheim~3D}
\newacronym{ICP}{ICP}{Iterative Closest Point}
\newacronym{ImgMA}{ImgMA}{Image Mesh Association}
\newacronym{LOD}{LOD}{Level Of Detail}
\newacronym{LUT}{LUT}{Lookup Table}
\newacronym[plural=MBBs, firstplural=Minimum Bounding Boxes (MBBs)]{MBB}{MBB}{Minimum Bounding Box}
\newacronym{ML}{ML}{Machine Learning}
\newacronym{MVS}{MVS}{Multi-View Stereo}
\newacronym{OA}{OA}{Overall Accuracy}
\newacronym[plural=PCs]{PC}{PC}{Point Cloud}
\newacronym{PCMA}{PCMA}{Point Cloud Mesh Association}
\newacronym{PCImgA}{PCImgA}{Point Cloud Image Association}
\newacronym{PSB}{PSB}{Princeton Shape Benchmark}
\newacronym{PS}{PS}{Persistent Scatterer}
\newacronym{RF}{RF}{Random Forest}
\newacronym{SACNN}{SACNN}{Structure-Aware Convolutional Neural Network}
\newacronym{SPG}{SPG}{Superpoint Graph}
\newacronym{UAV}{UAV}{Unmanned Airborne Vehicle}
\newacronym{V3D}{V3D}{Vaihingen~3D}

\begin{document}

\begin{frontmatter}

\title{Juggling With Representations: \\On the Information Transfer Between Imagery, Point~Clouds, and Meshes for Multi-Modal Semantics}

\author[ifp]{Dominik Laupheimer\corref{mycorrespondingauthor}} 
\ead{dominik.laupheimer@ifp.uni-stuttgart.de}
\author[ifp]{Norbert Haala}
\cortext[mycorrespondingauthor]{Corresponding author}
\address[ifp]{Institute for Photogrammetry, University of Stuttgart, Germany}

\begin{abstract}
The automatic semantic segmentation of the huge amount of acquired remote sensing data has become an important task in the last decade.
Images and \acp{PC} are fundamental data representations, particularly in urban mapping applications. 
Textured 3D meshes integrate both data representations geometrically by wiring the \ac{PC} and texturing the surface elements with available imagery. 
We present a mesh-centered holistic geometry-driven methodology that explicitly integrates entities of imagery, \ac{PC} and mesh. 
Due to its integrative character, we choose the mesh as the core representation that also helps to solve the visibility problem for points in imagery.
Utilizing the proposed multi-modal fusion as the backbone and considering the established entity relationships, we enable the sharing of information across the modalities imagery, \ac{PC} and mesh in a two-fold manner: (i)~feature transfer and (ii)~label transfer.
By these means, we achieve to enrich feature vectors to multi-modal feature vectors for each representation. 
Concurrently, we achieve to label all representations consistently while reducing the manual label effort to a single representation. 
Consequently, we facilitate to train machine learning algorithms and to semantically segment any of these data representations~\textendash~both in a multi-modal and single-modal sense.
The paper presents the association mechanism and the subsequent information transfer, which we believe are cornerstones for multi-modal scene analysis.
Furthermore, we discuss the preconditions and limitations of the presented approach in detail. 
We demonstrate the effectiveness of our methodology on the ISPRS 3D semantic labeling contest (Vaihingen~3D) and a proprietary data set (Hessigheim~3D). 
\end{abstract}

\begin{keyword}
Multi-Modality\sep Data Fusion\sep3D~Textured Mesh\sep 3D~Point Cloud\sep Imagery\sep Ground Truth \sep Semantic Segmentation
\end{keyword}

\end{frontmatter}


\glsresetall
\section{Introduction}
\label{sec:intro}
Over the years, data acquisition has become more redundant, more complete, faster, and denser~\textendash~spatially and temporally. 
Sensors such as cameras, LiDAR scanners, and RaDAR sensors guarantee multi-modal capturing of our world. 
Depending on the application and the desired mapping scale, the respective sensors are mounted on platforms such as satellites, airplanes, \acp{UAV}, or autonomous vehicles. 
In the domain of photogrammetry and remote sensing, particularly for urban mapping, data acquisition via imagery and \ac{ALS} is common.
Currently, data capture in urban areas at \acp{GSD} down to a few centimeters is becoming state of the art.
Traditionally, the airplane has been the platform of choice. 
However, more flexible and lightweight \acp{UAV} have grown in popularity in the past decade \citep{Haala2020}. 
\begin{figure}[htbp]
    \centering
    \begin{tabular}{lll}
        \includegraphics{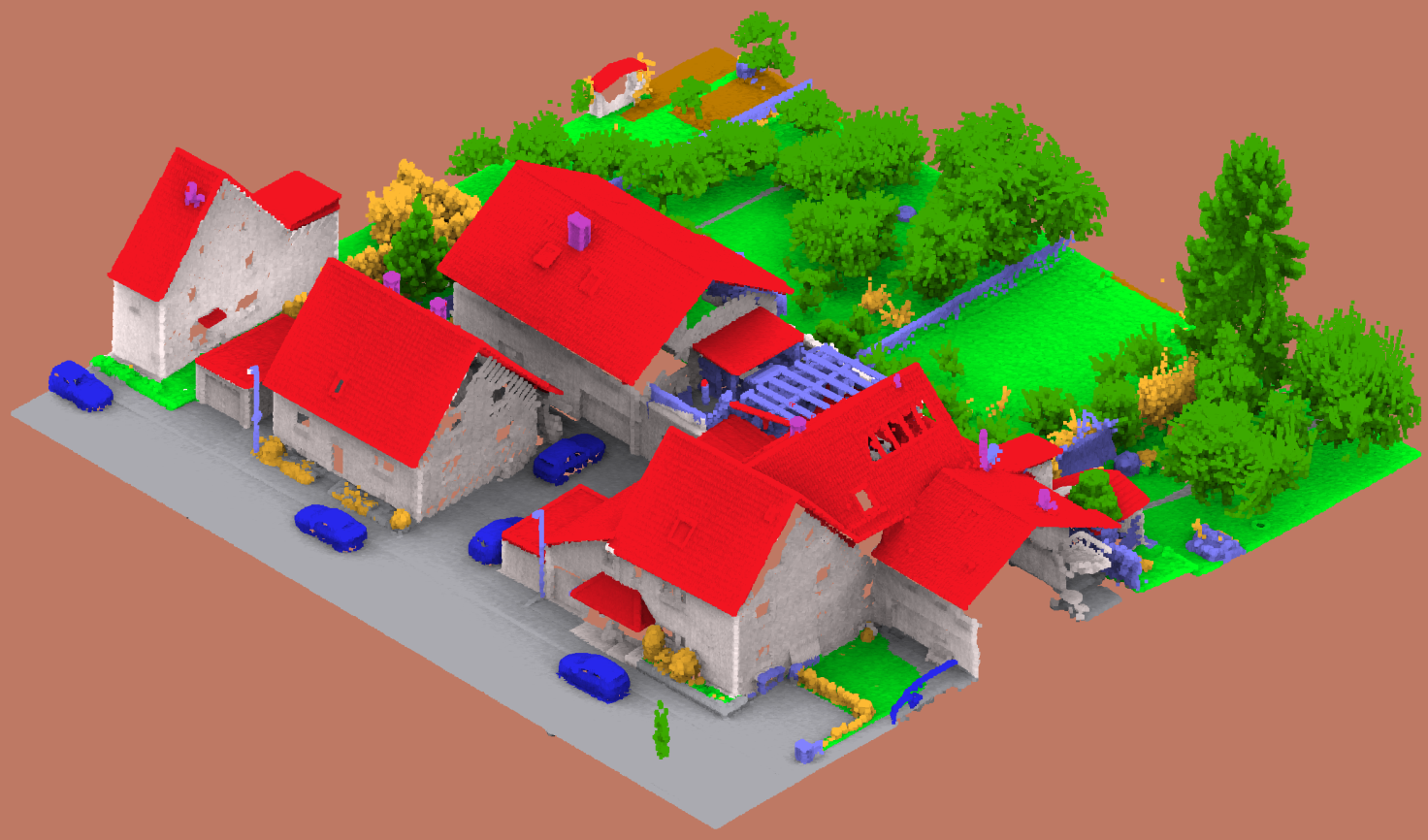}
        &
        \includegraphics{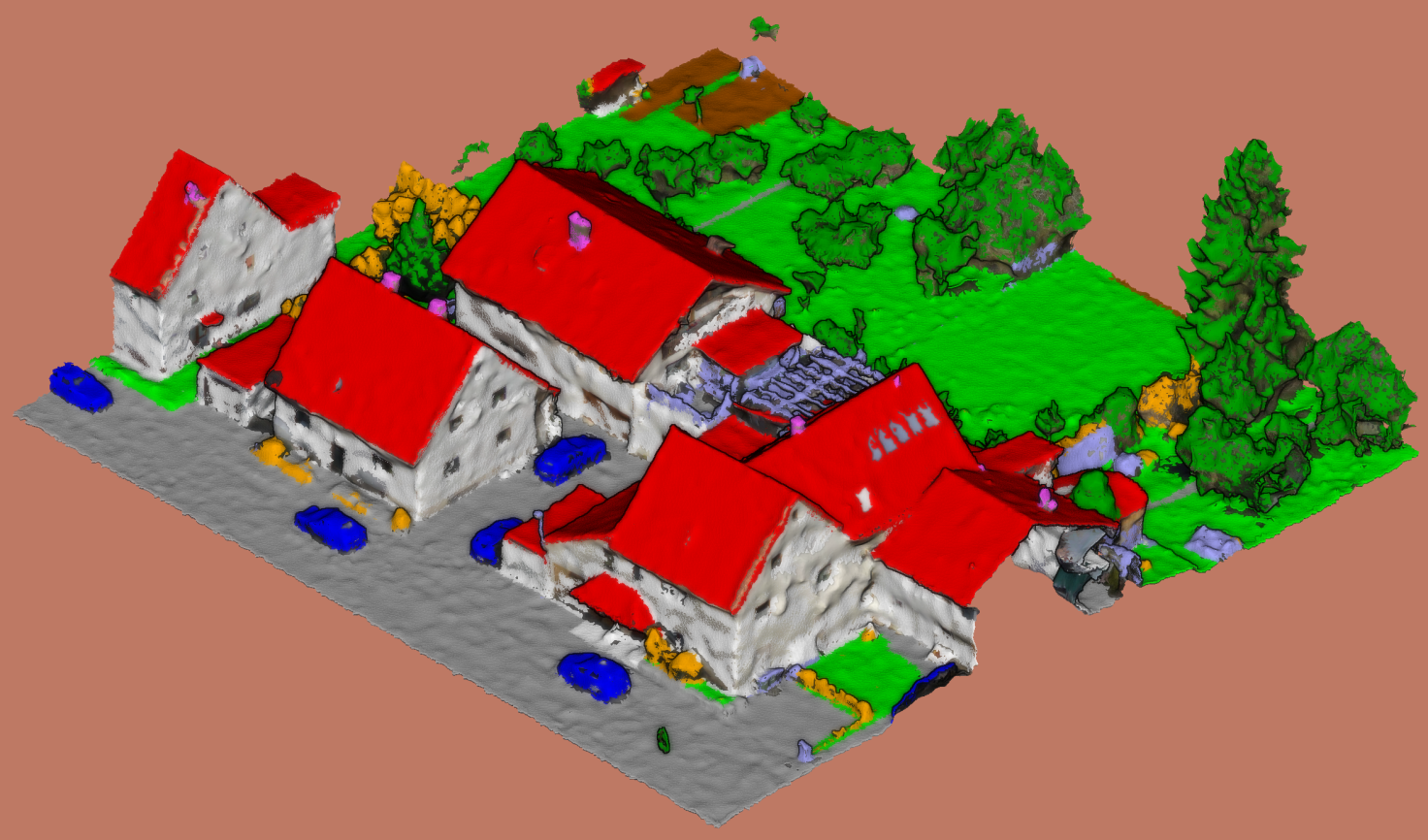}
        &
        \includegraphics{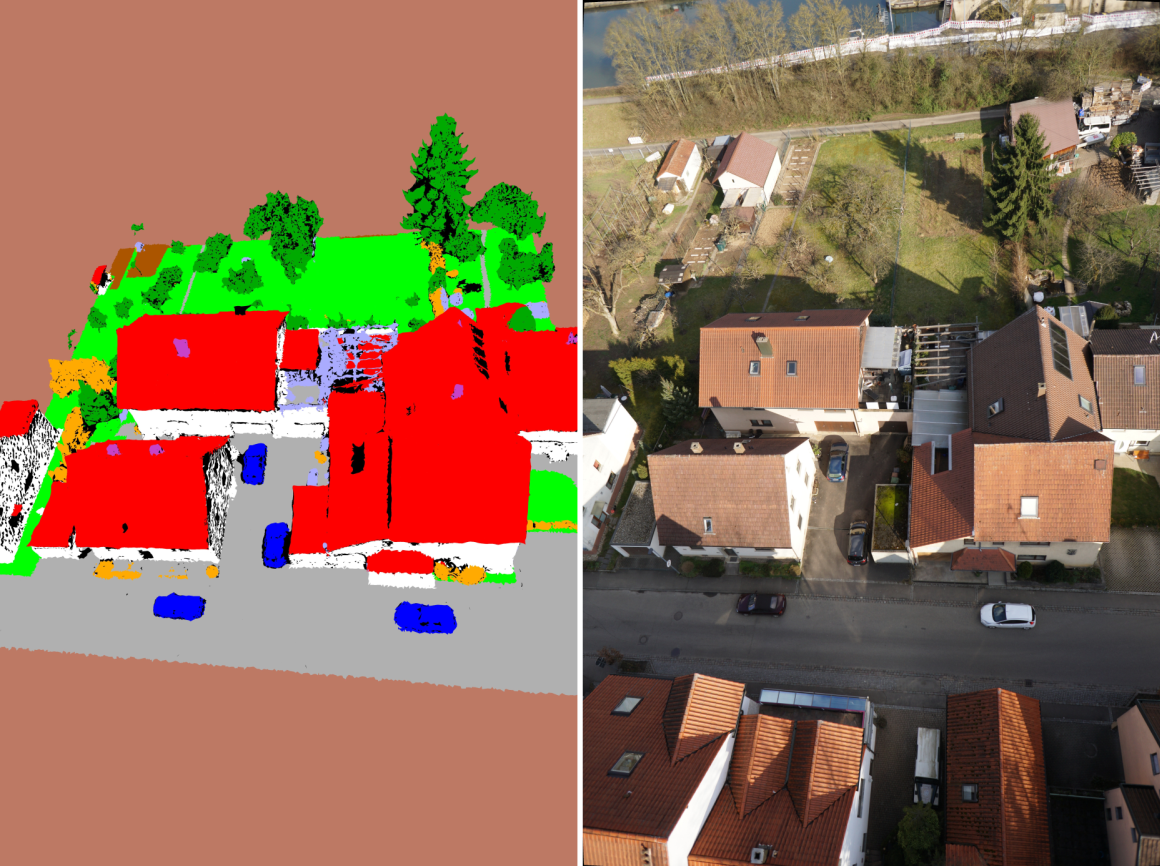}
    \end{tabular}
    \caption{
    Visualization of the proposed multi-modal data fusion by means of the enabled label transfer. The figure depicts the transferred annotations to the mesh (\textit{center}) and an oblique image (\textit{right}) as transferred from the respective manually labeled point cloud (a subset of Hessigheim~3D).
    Faces that cannot be linked to points are shown in textured fashion. Background or non-associated pixels respectively are colored in reddish. Pixels that are linked to an unlabeled face are depicted in black. The label scheme is given in~\autoref{fig:data_sets}.
    }
    \label{fig:juggler_alternative}
\end{figure}

Imagery is the fundamental photogrammetric data representation providing (multi-)spectral information.
Images project 3D real-world objects into 2D image space. 
By nature of the projection into grid-like pixel space, images suffer from occlusions, distortions, discretization, and the loss of the third dimension. 
However, with the help of automatic aerial triangulation, the intrinsic defects can be rectified and 3D reconstruction is possible. 
As a precondition to proper reconstruction, images have to provide unambiguous texture and capture each object point at least twice.
Derived 3D products of the \ac{MVS} pipeline are colored \acp{PC} and/or textured meshes, both mapping the surface of the captured region.

In contrast, due to the polar measurement principle, 3D \acp{PC} are the immediate \ac{ALS} output.
In comparison to \ac{MVS}, LiDAR scanning provides multi-target capability, and hence penetrates semi-transparent objects such as vegetation.  
Moreover, the polar measurement principle requires only a single measurement to map a 3D point. 
On the other hand, bare LiDAR points do not carry color/texture information like \acp{PC}/meshes as derived from imagery.
The accuracy of \ac{ALS} points depends on the accuracy of the trajectory. 
On the contrary, the accuracy of \ac{MVS} points is correlated with the \ac{GSD} which, theoretically, can be scaled arbitrarily. 
For a detailed comparison of these two capturing methodologies, we refer to \cite{Mandlburger2017}.

Initially, photogrammetry and laser scanning have been competitive systems with individual processing pipelines. 
However, at present, they are seen as complementary systems whose fusion results in more complete and better products. 
Nowadays, joint acquisition of photogrammetric and \ac{ALS} data is state of the art for airborne systems and starts to emerge even for \ac{UAV}-based systems \citep{Mandlburger2017,cramer2018monitoring}.
Recently, \cite{glira2019} proposed the hybrid orientation of \ac{ALS} \acp{PC} and aerial imagery, which improves the georeferencing accuracy of \ac{ALS} data by integrating stabilizing image block geometry into the strip adjustment. 
As a side product, the hybrid orientation enables a precise co-registration of imagery and LiDAR data.

Concerning the recent hybridization trend, from our point of view, enhancing 3D \acp{PC} to textured meshes may replace unstructured \acp{PC} as default representation for urban scenes in the future.
Intrinsically, meshes facilitate multi-modal data fusion by utilizing LiDAR points and \ac{MVS} points for the geometric reconstruction while leveraging high-resolution imagery for texturing (hybrid data storage).
Therefore, meshes are realistic-looking 3D maps of our real world and are easily understandable~\textendash~even for non-experts. 
Besides benefits for visualization, textured meshes have other favorable characteristics.
Whereas \acp{PC} are unordered sets of points, meshes are graphs consisting of vertices, edges, and faces that provide explicit adjacency information.
Meshes are less memory-consuming than \acp{PC} since meshing algorithms try to minimize the number of entities while reconstructing the maximum of detail. 
Before the meshing, \acp{PC} will be filtered in such a way that only geometrically relevant points are kept. 
This embraces noise filtering and filtering of points that can be approximated by the same face (e.g. points on planar surfaces). 
Furthermore, there will be geometric simplifications based on the desired level of detail and, as the case may be, due to 2.5D mesh geometry. 
By definition, meshes are surface descriptions that cannot handle multi-target capability like LiDAR \acp{PC}. 
This inevitably leads to a drop in entities to be stored. 
Moreover, the high-resolution texture information is stored in texture atlases avoiding redundant image content. 
Therefore, textured meshes provide geometric and textural information in a lightweight fashion.
Aside from these structural differences, georeferencing issues of imagery and  LiDAR data will cause discrepancies between imagery, \acp{PC}, and meshes, too.

Being a hybrid data storage, we believe that the mesh modality is ideally suited to foster multi-modal semantic analysis. 
To this end, we chose the mesh to be the core of the proposed multi-modal linking and transferring pipeline. 
The aims and objectives of the paper are postulated in~\autoref{subsec:aims_and_objectives}. 
In \autoref{sec:methodology}, we describe the entire methodology including the explicit entity linking and subsequent information transfer by deep-diving its building blocks.
Subsection~\ref{subsec:pcmeshassociation} describes the association of LiDAR data and the mesh in 3D space; \autoref{subsec:imgmeshassociation} outlines the association of imagery and the mesh. 
Both association mechanisms operate face-centered. 
Eventually, the combination of both links 3D points and pixels (cf.~\autoref{subsec:pcimgassociation}). 
Subsection~\ref{subsec:preconditions_limitations} discusses in detail prerequisites and particular challenges due to the mentioned (structural) discrepancies between the mesh, the \ac{PC}, and imagery.
In~\autoref{sec:results_discussion}, we demonstrate the proper working of the presented association mechanism on two real-world data sets \citep{V2D,cramer2018monitoring}.
Since \ac{GT} is not available across all modalities, a quantitative analysis of the proposed method is difficult. 
Therefore, we use the proposed label transfer to verify and showcase the proposed methodology.
Moreover, we report the best performing parameters for the association mechanism concerning the used imagery and \ac{ALS} data.
We briefly present the used data and the key parameters in~\autoref{sec:data}. 
Both data sets provide significantly different resolution and co-registration quality.

\subsection{Aims and Objectives}
\label{subsec:aims_and_objectives}
Our key contribution is the explicit linking of pixels, points, and faces to jointly leverage information from available data sources aiming at multi-modal semantics (cf.~\autoref{sec:methodology}).
Each face will be linked to several points and several pixels. 
In turn, points and pixels are linked while checking the visibility via the mesh. 
To the best of our knowledge, there is no other holistic approach that explicitly joins imagery, mesh, and LiDAR data.
The explicitly established connections on the entity-level are used to share information across modalities. 
Depending on the entity relationship, the information is aggregated prior to the transfer. 
The aggregation of features is achieved by calculating the median; label aggregation is achieved by majority vote (cf.~\autoref{tab:association_options}).  

In this study, we seem to focus on the label transfer since the effectiveness of the method can be shown better with labels than with features.
Furthermore, annotated \ac{GT} is of great importance for training supervised classifiers, particularly \ac{DL} approaches.
However, \ac{GT} generation is tedious, time-consuming, and expensive work wherefore real-world \ac{GT} availability is a rarity.
In particular, pixel-wise \ac{GT} generation is labor-intensive. 
Generally, there is a lack of \ac{GT} data sets that jointly provide \acp{PC} and oriented imagery (and in this way textured meshes). 
Therefore, available annotations are limited to a single representation and prevent exploiting the potential of multi-modal training.
This fact further motivates the necessity for the proposed approach and emphasizes its utility.
Our linking and transferring methodology facilitates the consistent labeling of various representations, given a manually annotated representation initially.
Hence, our method may help to overcome the imbalance of labeled entities among modalities. 
For instance, 3D data (\ac{PC} or mesh) can be projected into various images at a stroke (cf.~\autoref{fig:results_imagery}). 
Therefore, it minimizes the manual effort for \ac{GT} generation and helps considerably to foster modality-wise training of algorithms.

Our methodology is designed to process real-world data handling structural differences and co-registration issues.
To deal with the huge amount of redundant multi-modal data, the association operates in a tiled and parallelized fashion while aiming at a low memory footprint.
For the sake of good scientific practice, we critically reflect the preconditions and explore where the proposed approach might be limited (cf.~\autoref{subsec:preconditions_limitations}). 

To summarize, the proposed mesh-centered multi-modal entity linking serves as the backbone to share features and labels across entities. 
Representation-specific features and (manually generated) annotations can be shared at a stroke. 
Thus, the methodology allows the \textit{juggling with modalities} and injects great flexibility and versatility. 
The method enables the generation of multi-modal feature vectors and consistent annotation across modalities.
By these means, the proposed association mechanism fosters joint semantic analysis and consequently contributes to the completion of the hybrid processing pipeline.

\section{Related Work}
\label{sec:relatedwork}
The semantic segmentation of 3D data has become a standard task in the domain of photogrammetry and remote sensing. 
The increasing availability of simultaneously acquired airborne data with different acquisition methods calls for multi-modal fusion and scene analysis (cf.~subsections~\ref{subsec:relatedwork_datafusion} and~\ref{subsec:relatedwork_semseg}).
Generally, the semantic analysis deals with various representations such as imagery, voxels, \acp{PC}, and meshes. 
Regardless of modality, state-of-the-art \ac{ML} methods rely on a large amount of \ac{GT} data.
We briefly review available \ac{GT} of geospatial data in~\autoref{subsec:relatedwork_gt}.

\subsection{Multi-Modal Data Fusion}
\label{subsec:relatedwork_datafusion}
Due to complementary acquisition methods, multi-modal data acquisition has the potential to generate more complete and more detailed mapping products. 
Thereby, multi-modal products feature improved geometric reconstruction and semantic analysis. 
However, to the best of our knowledge, multi-modality is kept at a minimum and hence scratches only the surface.
For instance, the fusion of imagery and \ac{ALS} data on the point-level is commonly confined to the colorization of \ac{ALS} points \citep{V2D,cramer2018monitoring}.
To the best of our information, the explicit fusion of \ac{ALS} points and \ac{MVS} points and its contribution to semantics has not yet been investigated.
Possible reasons might be structural and georeferencing discrepancies across modalities as discussed in~\autoref{sec:intro} and the huge memory footprint as caused by redundant multi-modal capturing.
\cite{glira2019} propose a methodology to jointly orientate imagery and \ac{ALS} data which simplifies the fusion of the derived \acp{PC} as a side effect. 
Recently, there are software solutions that enable data fusion of multi-modal \acp{PC} and refine the fusion on the mesh-level. 
For instance, software SURE by nFrames \citep{rothermel2012sure} produces meshes as generated from LiDAR and \ac{MVS} dealing with orientation discrepancies of few \acp{GSD}.

As outlined in \autoref{subsec:relatedwork_semseg}, the majority of works for semantic interpretation involves only one modality in the narrow sense.
In most cases, multi-modality is a means to an end that allows abusing annotated data and well-performing classifiers of another modality.
To give an example, the well-established and fast semantic segmentation of images is mostly abused as a proxy to 3D scene analysis \citep{Boulch2017,Lawin2017,He2013,Su2015,Kalogerakis_labelMeshes}.
Theoretically, any quantity can be projected into image space adding another channel to the image. 
In practice, the curse of dimensionality prevents the projection of an arbitrary number of quantities.
\cite{Peters2019} highlight issues of associating \acp{PC} and imagery, particularly time-shifts and occlusions. 
To by-pass the occlusion problem, they approximate the 3D surface by voxelization of the \ac{PC}.

Our work differs from existing works since it explicitly aims at a holistic multi-modal data fusion of imagery, \acp{PC} and meshes. 
Thereby, the mesh acts as core modality to solve the occlusion problem.
The subsequent information transfer shares features and labels with all modalities.
The association of an \ac{ALS} \ac{PC} and a challenging 2.5D mesh is already described in \cite{Laupheimer2020_ISPRS}. 
In the current work, we improve the implementation to cope with 3D meshes with a significantly larger memory footprint than 2.5D meshes.
Moreover, we extend the association mechanism to image space  (cf.~\autoref{subsec:imgmeshassociation} and~\autoref{subsec:pcimgassociation}) and enable information transfer in arbitrary directions (cf.~\autoref{tab:association_options}).

\subsection{Semantic Segmentation of 3D Data}
\label{subsec:relatedwork_semseg}
\ac{DL} methods, particularly \acp{CNN}, are state of the art for semantic segmentation in image space \citep{Garcia-Garcia2017,Minaee2020}.
Therefore, it seems reasonable to apply well-established \ac{DL} methods of the image space to \acp{PC}.
However, the unstructured nature of 3D \acp{PC} prevents to apply \acp{CNN} directly to them. 
To overcome the non-Euclidean design, \acp{PC} are commonly structured into grid-like 3D or 2D representations by voxelization or multi-view rendering respectively.
Several works voxelize the \ac{PC} and train a supervised classifier. The predicted labels for the voxels will be transferred to all contained points \citep{hackel_pc_segementation,Huang2016}. 
Voxelization comes along with memory overhead. 
Therefore, much effort is put into networks that use sparse 3D convolutions \citep{Graham2018}.
This approach has been successfully applied to urban \acp{PC} \citep{Schmohl2019}.
Detouring via image space, multi-view approaches leverage well-performing semantic image segmentation methods. The per-pixel predictions are back-projected to 3D space \citep{Boulch2017,Lawin2017}.
To give an example, \cite{He2013} segment stereo images semantically, create the \ac{MVS} \ac{PC}, and back-project the 2D semantic segmentation results to the \ac{PC}.
The grid-like proxy enables the use of \acp{CNN} but comes along with information loss due to discretization, occlusions, and projection. 

The rise of PointNet and its hierarchical successor PointNet++ constitutes a milestone in semantic \ac{PC} segmentation since they operate directly on unstructured 3D \acp{PC} \citep{qi2017pointnet, qi2017pointnet++}. \cite{Winiwarter2019} successfully applied PointNet++ to geospatial \acp{PC}.
The gist of PointNet is to use a symmetric function during encoding to be independent of set permutation. 
The entire \ac{PC} is encoded by a global feature vector, which is attached to each encoded per-point feature vector. 
Operating only on a global scale, PointNet misses local context.
Its extension PointNet++ hierarchically applies PointNets to the iteratively subsampled \ac{PC} and, hence, operates on several scales. 
This procedure mimics hierarchical feature learning with increased contextual information similar to \acp{CNN} in image space.
Likewise, \cite{Boulch2019} introduces continuous convolutional kernels that can be applied directly to \acp{PC}. 
\cite{Griffiths2019a} review the current state-of-the-art \ac{DL} architectures for processing 3D data.
\cite{Xie2020} review semantic \ac{PC} segmentation comparing \ac{DL} and traditional \ac{ML} approaches. 
While \ac{DL} approaches do not require handcrafted features, they rely on a large amount of training data.
In contrast, traditional \ac{ML} depends on handcrafted features and therefore provides better interpretability.
\cite{WeinmannEtAl2015} calculate and select features based on various vicinities and subsequently perform a semantic segmentation with \ac{RF}. 
To avoid noisy predictions, \cite{Landrieu2017b} extend the previous pipeline by structured regularization, a graph-based contextual strategy.
Likewise, \cite{NIEMEYER2014} avoid noisy results utilizing CRF-based methods as statistical context models. 
\cite{Vosselman2017} first segment the data and subsequentially perform the semantic segmentation. 

\cite{Ahmed2018} show advances of \ac{DL} on different 3D data representations. 
They discuss representation-specific challenges and highlight differences between Euclidean and non-Euclidean data.
The emerging field of geometric \ac{DL} extends basic \ac{DL} operations to non-Euclidean domains such as graphs and manifolds in order to use topological information \citep{Bronstein2016}.
\acp{PC} do not provide topological information per se. 
Therefore, \cite{Landrieu2017} organize \acp{PC} in \acp{SPG}. 
\cite{AliKhan2020} transform \acp{PC} to an undirected symmetrically weighted graph encoding the spatial neighborhood and apply a Graph Convolutional Network.
\cite{Chang2018} propose the \ac{SACNN} that uses generalized filters, which aggregate local inputs of different learnable topological structures. 
By that, \glspl{SACNN} work with both Euclidean and non-Euclidean data. 
To summarize, the adaption of (geometric) \ac{DL} methods contributed to substantial progress in the field of semantic \ac{PC} segmentation in the last decade. 

On the contrary, mesh interpretation has hardly been explored by the community of photogrammetry and remote sensing although recent years show increasing interest in meshed 3D models - particularly, for applications like smart city models \citep{Boussaha2018}. 
In comparison, meshes are a default data representation in the domain of computer vision. 
However, that community typically deals with small-scale (indoor) data sets \citep{Kalogerakis_labelMeshes}.
In contrast to photogrammetric meshes, texture is not an inherent characteristic of these meshed models.
By analogy to semantic \ac{PC} segmentation, common approaches for semantic mesh segmentation make a circuit to 2D image space to take advantage of image-based \ac{DL} methods. 
Those approaches render 2D views of the 3D scene, learn the segmentation for different views, and finally, back-project the segmented 2D images onto the 3D surface \citep{Su2015,Kalogerakis_labelMeshes}. 
\cite{wu20153d} voxelize the mesh and apply a convolutional deep belief network.
\cite{qiao2019laplaciannet} propose a geometric \ac{DL} approach that encodes the mesh connectivity using Laplacian spectral analysis and aggregates global information via mesh pooling blocks. 
MeshCNN mimics traditional \ac{CNN} convolution and pooling operations \citep{Hanocka2019}.
The specialized layers operate on the edges and leverage the intrinsic topological information.
\cite{Schult20CVPR} propose the DualConvMesh-Net that combines geodesic and Euclidean convolutions on 3D meshes. 
Geodesic convolutions utilize the underlying mesh structure and help to separate spatially adjacent but disconnected surfaces. 
In contrast, Euclidean convolutions establish connections between nearby disconnected surfaces. 

Notwithstanding, semantic segmentation of real-world large-scale meshes is a mostly overlooked topic.
\cite{Rouhani2017} gather faces of a \ac{MVS} mesh into so-called superfacets and train a \ac{RF} using geometric and photometric features.
\cite{tutzauer2019} utilize a \ac{DL} approach by training a multi-branch 1D \ac{CNN} with contextual features and compare the achieved results to a \ac{RF}. 
They show that color information is beneficial for semantic mesh segmentation.
More precisely, \cite{Laupheimer2020_DGPF} attest that per-face color information (i.e. texture) outperforms per-vertex color information (e.g. colored \ac{PC}) by evaluation of several radiometric feature qualities.
However, they also show the inherent limitations of texture due to occlusions, absence of imagery, and the quality of the geometric reconstruction.

\subsection{Ground Truth Availability}
\label{subsec:relatedwork_gt}
\cite{Garcia-Garcia2017} and \cite{Minaee2020} review available \ac{GT} data in image space, 2.5D and 3D space. 
Annotated imagery often aims at the pure semantic segmentation, wherefore orientation information is not provided \citep{Lambert2020}.
The ISPRS 2D Semantic Labeling Contest provides manually annotated orthophotos of Vaihingen and Potsdam \citep{V2D}.
\cite{Griffiths2019a} list available \ac{GT} data sets for RGB\hbox{-}D, multi-view, volumetric, and fully end-to-end architecture designs as acquired by various platforms.
\cite{Xie2020} review publicly available annotated \acp{PC} and discuss their shortcomings.

The computer vision community provides annotated mesh data for indoor scenes \citep{Armeni2017,Hua2016, Dai2017} or for single objects \citep{shilane2004princeton}.
However, to the best of our knowledge, there are no labeled meshed models that cover urban scenes.
In contrast, there are many available labeled urban data sets for 3D \acp{PC} provided by the community of photogrammetry and remote sensing \citep{Zolanvari2019, Wichmann2018, NIEMEYER2014,Hackel2017}.

The rise of data-hungry \ac{DL} methods demands efficient strategies for \ac{GT} generation. 
Synthetically generated \ac{GT} such as provided by \cite{Griffiths2019b} boost the generation process per se. 
However, purely synthetic data is limited by its diversity.
\cite{Koelle2020} exploit crowdsourcing and active learning to minimize manual labeling effort.  
\cite{ramirez2019shooting} present a virtual reality tool that gamifies the manual labeling of meshes and \acp{PC}.

Our proposed methodology is able to derive consistently labeled meshes and imagery from publicly available annotated real-world \ac{PC} data and vice versa (provided that the necessary data is available and oriented, cf.~\autoref{sec:results_discussion}). 
To the best of our knowledge, yet, there is no data set that provides consistently labeled modalities. 
The proposed labeling tool has the potential to accelerate multi-modal \ac{GT} generation and consequently multi-modal semantic analysis.

\section{Data}
\label{sec:data}
To demonstrate the effectiveness of our association mechanism, we utilize the publicly available ISPRS benchmark data set \ac{V3D} and a proprietary data set which will be made publicly available in mid 2021 \citep{Cramer2010,cramer2018monitoring}. 
The original purpose of the proprietary data set aims at the deformation monitoring of the ship lock and its surrounding in Hessigheim, Germany. 
Thus, challenging water surfaces are part of the acquired data.
We refer to this data set as \ac{H3D}.
Although being already captured in 2008, \ac{V3D} may still be representative of large-scale country-wide mapping.
On the contrary, \ac{H3D} is an example of small-scale mapping applications with high-resolution imagery and LiDAR data. 

In both cases, imagery and \ac{ALS} data have been acquired from airborne platforms. 
Whereas \ac{V3D} data is captured from airplane, \ac{H3D} data is captured from \acp{UAV}.
\ac{V3D} data has been acquired asynchronously. 
The time-shift between nadir imagery (GSD\,=\,\SI{8}{cm}) and \ac{ALS} acquisition (\SIrange[range-units=single, range-phrase=--]{4}{8}{points/m^2}) is several weeks.
\ac{H3D} provides two sets of oriented imagery: oblique and nadir.
Oblique imagery (GSD\,=\,\SI{2.5}{cm}) has been acquired simultaneously along with \ac{ALS} data (\SIrange[range-units=single, range-phrase=--]{400}{800}{points/m^2}) from the same \ac{UAV}. 
Nadir images (GSD\,=\,\SI{3.7}{mm}) have been acquired from another \ac{UAV} with a time-shift of several hours to one day.
Accordingly, the number of entities and the memory footprint is higher for \ac{H3D}.
\autoref{tab:data} lists key parameters of both data sets relevant for the underlying study.
\begin{table}[htbp]
    \centering
    \footnotesize
    \begin{tabular}{P{2.5cm}P{3cm}P{2.5cm}P{2.5cm}}
        \toprule
        \normalsize Data Set & 
        \normalsize Imagery & 
        \normalsize PC & 
        \normalsize Mesh 
        \\
        \toprule
        \minitab{
        \ac{V3D} \\ 
        \SI{870}{m}~$\times$~\SI{700}{m} 
        } 
        & 
        \minitab{
        $\text{GSD}_{\text{nadir}} = \SI{8}{cm}$ \\
        15 images \\ 
        @ \SI{14430}{px} $\times$ \SI{9420}{px}
        }
        & 
        \minitab{
        \SIrange[range-units=single, range-phrase=--]{4}{8}{points/m^2} \\
        \SI{1.2}{M} points
        }
        &
        \minitab{
         16 tiles \\ 
         \SI{3.3}{M} faces \\
         source: \ac{MVS}$_{\text{nadir}}$
        }
        \\
        \midrule
        \minitab{
        \ac{H3D} \\ 
        \SI{670}{m}~$\times$~\SI{255}{m}
        }
        &
        \minitab{
        $\text{GSD}_{\text{oblique}} = \SI{2.5}{cm}$ \\ 
        1979 images \\
        @ \SI{6000}{px} $\times$ \SI{4000}{px} \\
        \cline{1-1}$\text{GSD}_{\text{nadir}} = \SI{3.7}{mm}$ \\ 
        524 images \\
        @ \SI{11608}{px} $\times$ \SI{8708}{px}   
        }
        &
        \minitab{
        \SIrange[range-units=single, range-phrase=--]{400}{800}{points/m^2} \\
        $\SI{138.1}{M}$ points  
        }
        &
        \minitab{
        94 tiles \\ 
        $\SI{24.6}{M}$ faces\\
        source: \\
        \ac{MVS}$_{\text{oblique}}$ $+$ \ac{ALS}
        }
        \\
		\bottomrule
	\end{tabular}
    \caption{\ac{V3D} and~\ac{H3D} properties. The project area is given by the mesh tiles that intersect with the labeled LiDAR cloud. The face count is adapted to the overlapping LiDAR cloud. 
    }
    \label{tab:data}
\end{table}

For both data sets, we generate textured and tiled meshes with SURE 4.0.2 from nFrames. 
We set tile sizes to $\SI{175}{m} \times \SI{175}{m}$ (\ac{V3D}) and $\SI{50}{m} \times \SI{50}{m}$ (\ac{H3D}) respectively. 
The chosen tile sizes empirically showed to be a good compromise between fast tile-wise processing and small tile count. 
For H3D, we generate a hybrid textured mesh by fusing the simultaneously acquired \ac{ALS} data and oblique imagery. 
Utilizing oblique imagery ensures proper texturing of vertical faces such as facades. 
In contrast, we generate a purely photogrammetric mesh for \ac{V3D} since the time-shift of imagery and \ac{ALS} data is roughly one month. 
For this reason, the geometric and radiometric quality of the \ac{H3D} mesh outperforms the \ac{V3D} mesh. 
However, since the \ac{V3D} mesh is purely photogrammetric, the relative orientation of imagery and mesh fits perfectly. 
We determine the shifts between mesh and \ac{PC} data with the \ac{ICP} algorithm. 
We do not apply the determined shifts to prove the effectiveness of our association methodology. 
Moreover, \ac{ICP} does not solve the co-registration problem entirely. 
The \ac{V3D} \ac{ALS} data is shifted against the \ac{MVS} mesh by 
\mbox{$\Delta X=\SI{31}{cm}$}, 
\mbox{$\Delta Y=\SI{-9}{cm}$}, 
\mbox{$\Delta Z=\SI{9}{cm}$}.  
Rephrased, the co-registration of imagery and \ac{ALS} data differs significantly.
The \ac{H3D} \ac{ALS} data is shifted against the mesh (as generated from LiDAR and \ac{MVS} points) by 
\mbox{$\Delta X=\SI{8}{mm}$}, 
\mbox{$\Delta Y=\SI{10}{mm}$}, 
\mbox{$\Delta Z=\SI{6}{mm}$}. 
The significantly different data sets featuring co-registration issues in 3D space are adequate to showcase the robustness and flexibility of our implementation.

\begin{figure}[htbp]
    \centering
    \begin{tabular}{cc}
        \includegraphics[width=0.568\columnwidth]{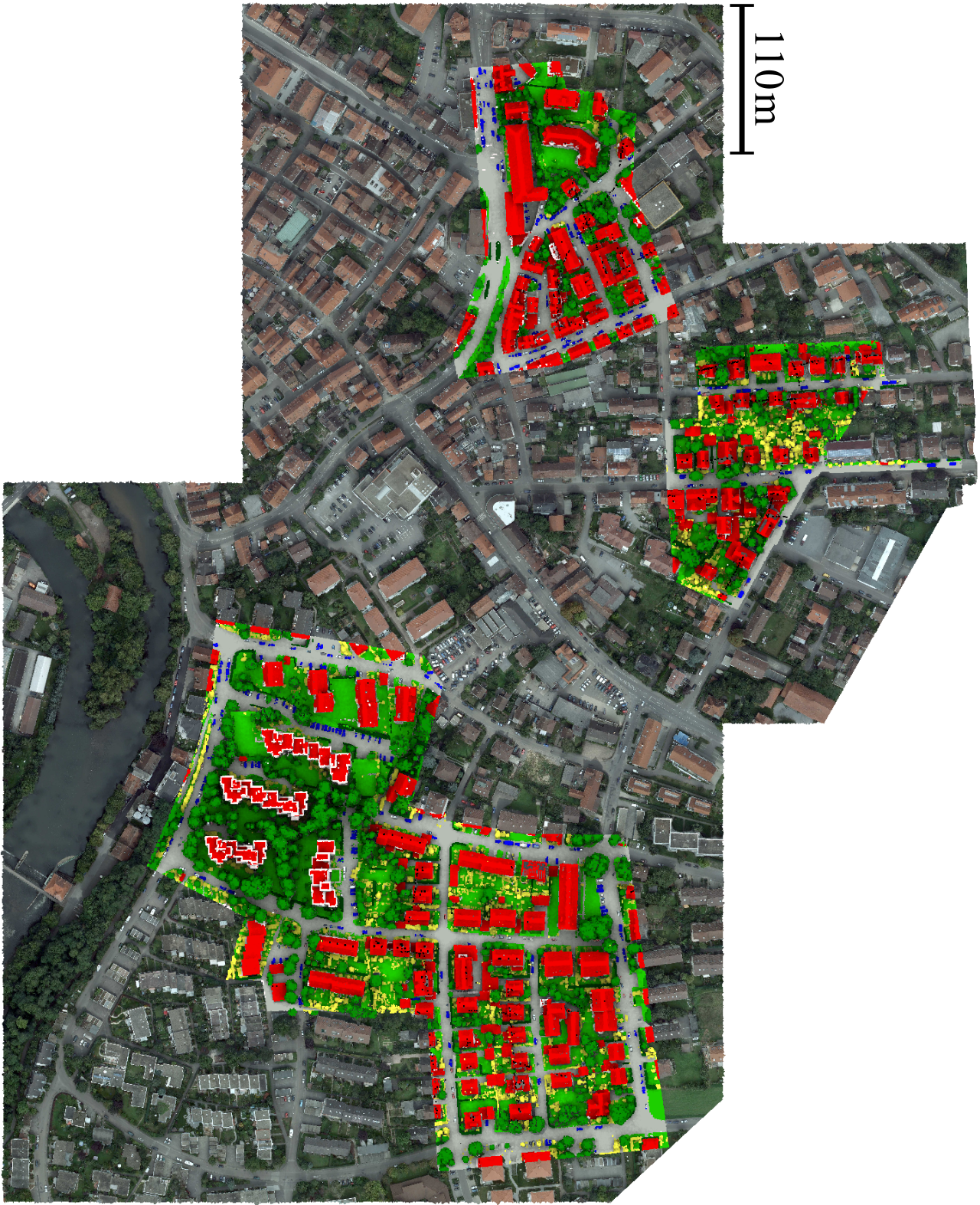} & 
        \includegraphics[width=0.285\columnwidth]{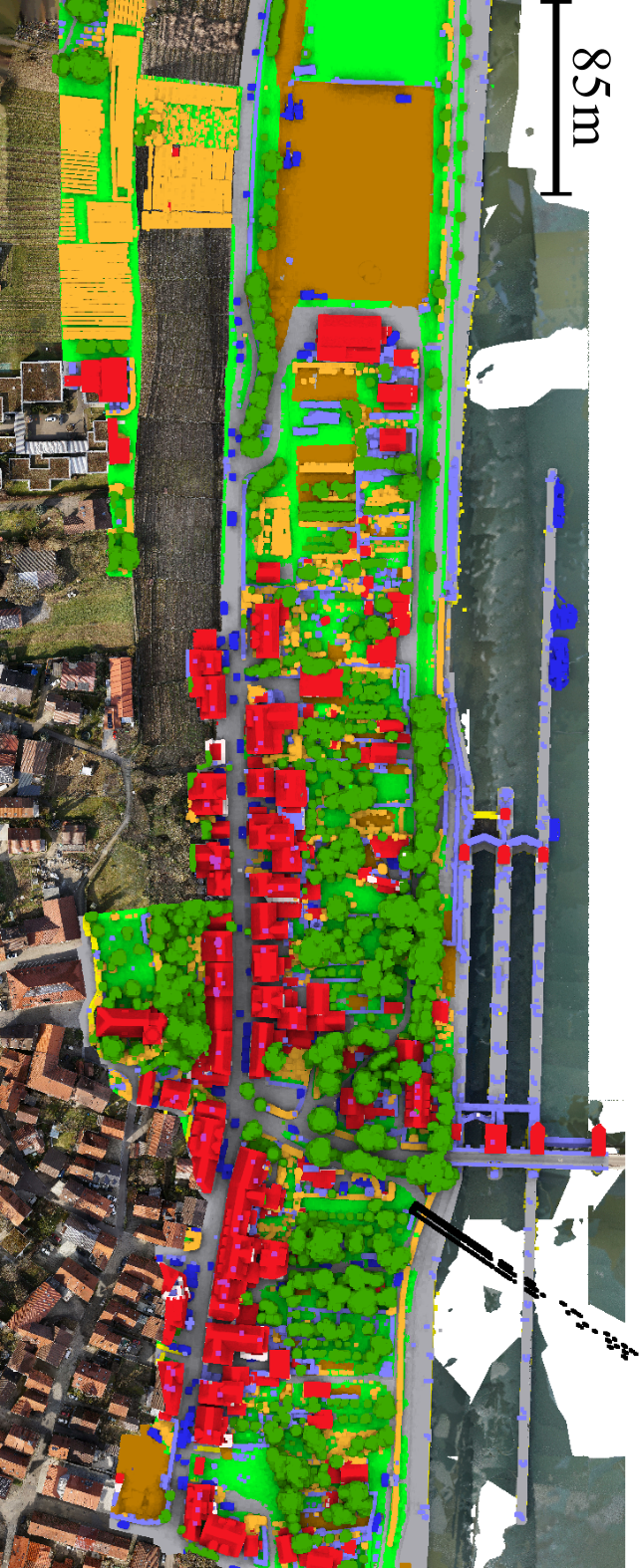}
    \end{tabular}
    \caption{
    Top views of \ac{V3D} (\textit{left}) and \ac{H3D} (\textit{right}) depicting the annotated LiDAR \ac{PC} and the respective overlapping mesh tiles in textured fashion.
    The annotated \ac{ALS} data is color-coded utilizing the following label schemes. 
    \newline
    \ac{V3D}: \newline
    \textit{Power Line} (black), \textit{Low Vegetation} (light green), \textit{Impervious Surface} (gray), \textit{Car} (blue), \textit{Fence/Hedge} (yellow), \textit{Roof} (red), \textit{Facade} (white), \textit{Shrub} (dark green), and \textit{Tree} (green).
    \newline
    \ac{H3D}:\newline 
    \textit{Power Line} (black), \textit{Low Vegetation} (light green), \textit{Impervious Surface} (gray), \textit{Vehicle} (blue), \textit{Urban Furniture} (lilac), \textit{Roof} (red), \textit{Facade} (white), \textit{Shrub/Hedge} (orange), \textit{Tree} (green), \textit{Open Soil/Gravel} (brown), \textit{Vertical Face} (yellow), \textit{Chimney/Antenna} (magenta).
    }
    \label{fig:data_sets}
\end{figure}

Both data sets carry manual annotations for the LiDAR cloud \citep{NIEMEYER2014,KoelleEtAl2019}.
The label scheme of \ac{H3D} is oriented towards the \ac{V3D} label scheme but is more fine-grained. 
Furthermore, it has been manually enhanced by class \textit{Chimney/Antenna} \citep{Laupheimer2020_DGPF}.
\autoref{fig:data_sets} shows the union of textured mesh tiles that overlap with the labeled LiDAR cloud for both data sets. 
The label schemes and respective color-codings are given in the figure caption.

\section{Methodology}
\label{sec:methodology}
We aim for a holistic explicit linking of the common data representations in the domain of photogrammetry and remote sensing: imagery, \ac{PC}, and mesh. 
The backbone of the proposed association methodology consists of two geometry-driven parts:
(a)~\ac{PCMA} which links faces and points (cf.~\autoref{subsec:pcmeshassociation}) and (b)~\ac{ImgMA} which links faces and pixels across images (cf.~\autoref{subsec:imgmeshassociation}). 
Coupling both association mechanisms yields to (c)~\ac{PCImgA} (cf.~\autoref{subsec:pcimgassociation}). 
The \ac{PCImgA} establishes a connection between points and imagery via the mesh as a mediator. 
Point visibility is implicitly given through the mesh.
\autoref{tab:association_options} illustrates the total association mechanism with iconic pictograms. 
\begin{table}[htbp]
\resizebox{\textwidth}{!}{
    \centering
    \footnotesize
    \begin{tabular}{cccc}
        \toprule
        \multicolumn{4}{c}{} \\
        \multicolumn{4}{c}{
            \centering
            \includegraphics[width=1.0\linewidth]{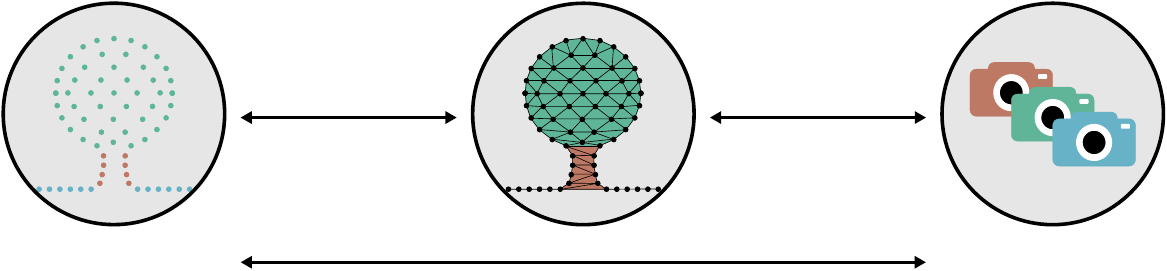}
            \put (-299.5, 80) {Point Cloud Mesh Association}
            \put (-150, 80) {Image Mesh Association}
            \put (-229.5,-10) {Point Cloud Image Association}
            \put (-255, 48) {$n_\mathit{pts}:1$}
            \put (-127, 48) {$1:n_\mathit{px}:n_\mathit{img}$}
            \vspace{0.5cm}
        } \\
        \toprule
        \toprule
        \multicolumn{4}{l}{(a) \acrfull{PCMA}}\\
        \bottomrule
        &
        Mesh $\mapsto$ PC ($1:n_\mathit{pts}$) & 
        PC $\mapsto$ Mesh ($n_\mathit{pts}:1$) & 
        \multirow{4}{*}{
            \centering
            \includegraphics[width=0.10\linewidth]{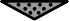}
        }\\
        \cmidrule[0.5pt](l{0.5em} r{0.5em}){2-2}
        \cmidrule[0.5pt](l{0.5em} r{0.5em}){3-3}
        Feature Transfer &
        Copy Value & 
        Median Aggregation \\
        Label Transfer &
        Copy Value & 
        Majority Vote & \\
        & & \\
        \toprule
        \multicolumn{4}{l}{(b) \acrfull{ImgMA}}\\
        \bottomrule
		& 
		Mesh $\mapsto$ Img ($1:n_\mathit{px}:n_\mathit{img}$) & 
        Img $\mapsto$ Mesh ($n_\mathit{img}:n_\mathit{px}:1$) &
        \multirow{4}{*}{
            \centering
            \includegraphics[width=0.25\linewidth]{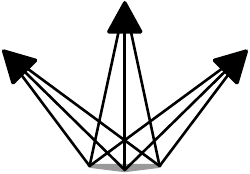}
        }\\
        \cmidrule[0.5pt](l{0.5em} r{0.5em}){2-2}
        \cmidrule[0.5pt](l{0.5em} r{0.5em}){3-3}
        Feature Transfer &
        Copy Value & 
        Median Aggregation &\\
        Label Transfer &
        Copy Value & 
        Majority Vote & \\
        & & &\\
        \toprule
        \multicolumn{4}{l}{(c) \acrfull{PCImgA}}\\
        \bottomrule
		&
		PC $\mapsto$ Img & 
        Img $\mapsto$ PC  &
        \multirow{4}{*}{
            \centering
            \includegraphics[width=0.25\linewidth]{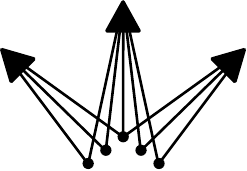}
        }\\
        \cmidrule[0.5pt](l{0.5em} r{0.5em}){2-2}
        \cmidrule[0.5pt](l{0.5em} r{0.5em}){3-3}
        Feature Transfer &
        Median Aggregation & 
        Median Aggregation & \\
        Label Transfer &
        Majority Vote & 
        Majority Vote & \\
        & & & \\
		\bottomrule
	\end{tabular}
	} 
    \caption{Overview of the proposed method. For each association mechanism, the transfer operations are given in dependence of the information type (feature or label) and the transfer direction. 
    The pictograms on the right depict the linking of the respective entities.
    \ac{PCImgA} provides two association modes: implicit and explicit linking (cf.~\autoref{subsec:pcimgassociation}). 
    (a), (b) and the implicit version of (c) are face-centered. 
    The relationship of implicit \ac{PCImgA} is described by $n_\mathit{pts}:1:n_\mathit{px}:n_\mathit{img}$. 
    The explicit version is pixel-centered (PC~$\mapsto$~Img: $n_\mathit{pts}:1:n_\mathit{img}$) or point-centered (Img~$\mapsto$~PC: $n_\mathit{img}:n_\mathit{px}:1$).
    }
    \label{tab:association_options}
\end{table}

The established connections between the entities across the distinct representations enable an information transfer that allows features and labels to be shared arbitrarily.
\autoref{tab:association_options} compactly lists the information transfer operations depending on information type (feature or label) and transfer direction for each part of the entire association mechanism.
Concerning the scalability of the proposed multi-modal association approach, we process data tile-wise in a parallelized fashion while keeping the memory footprint low.

\subsection{Point Cloud Mesh Association (PCMA)}
\label{subsec:pcmeshassociation}
\begin{figure}[htbp]
    \centering
    \includegraphics[width=0.675\linewidth]{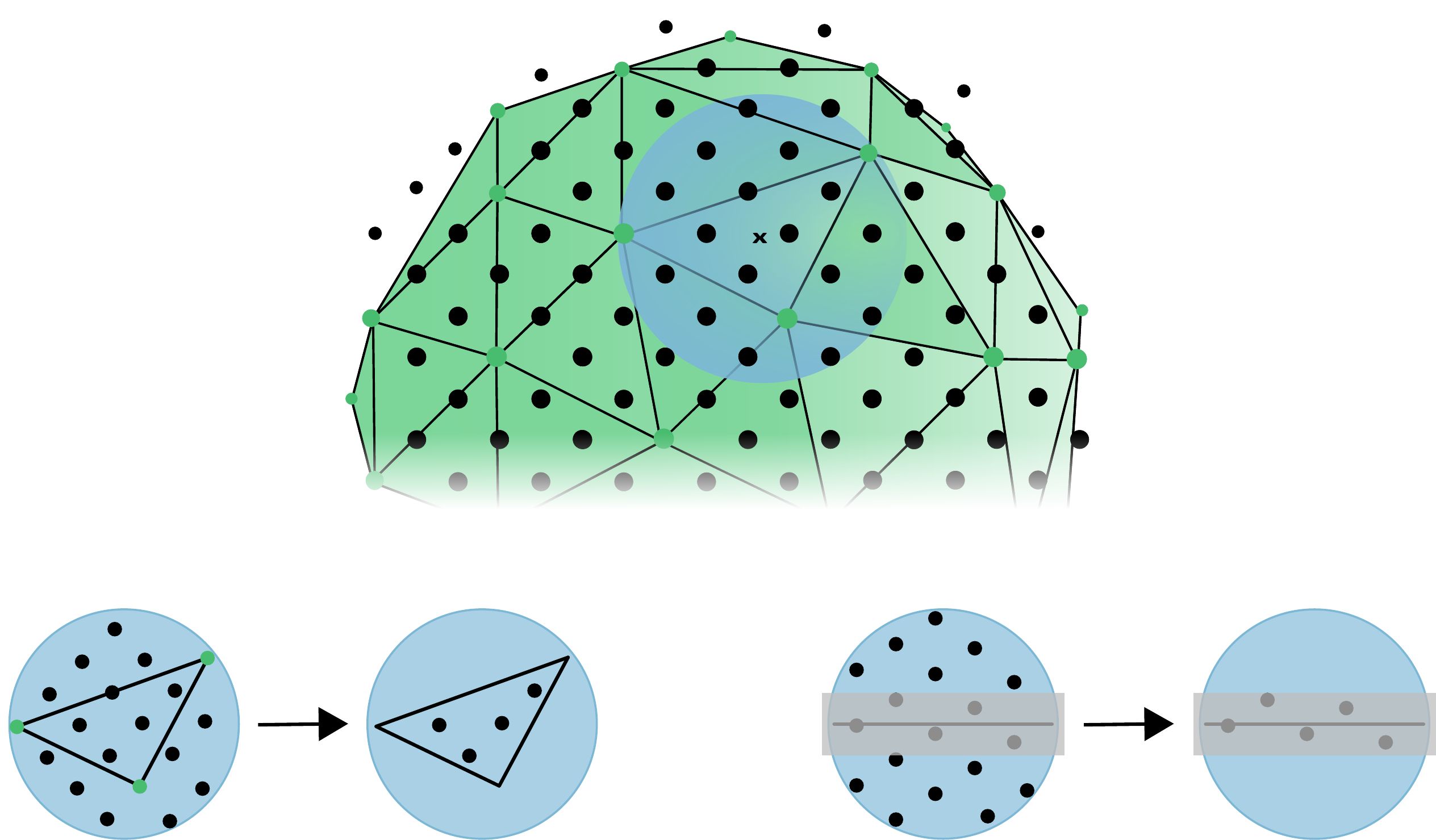}
    \put (-122.5, 43) {(i)}
    \put (-191.5, -4) {(ii)}
    \put (-58, -4) {(iii)}
    \caption{
    Steps (i) - (iii) of the \ac{PCMA}.
    \textit{(i)}: Clipping of the \ac{PC} (black dots) to the vicinity (blue sphere) of the considered face. Its \ac{COG} is marked with a black cross. 
    The mesh surface and its vertices are depicted in green.
    \textit{(ii)}: Filtering of \textit{out-of-face points} based on the clipping result (orthogonal view concerning the face surface).
    \textit{(iii)}: Filtering of \textit{off-the-face points} (side view with respect to the face). 
    The face is depicted as a black line. 
    The threshold band is marked in gray.}
    \label{fig:methodology_association_pc_mesh}
\end{figure}

The \ac{PCMA} explicitly links faces and points in a face-centered geometry-driven approach. 
Each face (represented by its \ac{COG}) is assigned with $n_\mathit{pts}$~points that represent the same surface by following three steps:
(i) clipping of the \ac{PC} to a spherical vicinity of the \ac{COG},
(ii) filtering of \textit{out-of-face points}, 
and (iii) filtering of \textit{off-the-face points} (\autoref{fig:methodology_association_pc_mesh}).
\textit{Out-of-face points} are not enclosed by the face borders when projected orthogonally onto the face plane.
\textit{Off-the-face points} do not coincide with the face plane, i.e. they are below or above the face surface. 
A manually set threshold~$\theta$ decides whether a point coincides with a face or not.
Both point types are not mutually exclusive and exist due to the simplification during the meshing, the representation type differences as discussed in~\autoref{sec:intro}, and geometry differences (e.g. in case of 2.5D mesh geometry or due to asynchronous data acquisition).

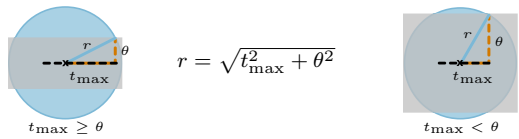
\begin{figure}[htbp]
    \centering
    \includegraphics[width=0.125\linewidth]{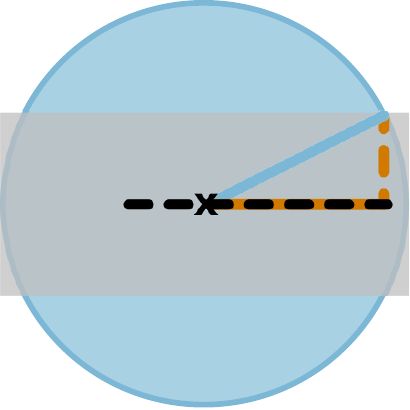}
    \hspace{3.5cm}
    \includegraphics[width=0.125\linewidth]{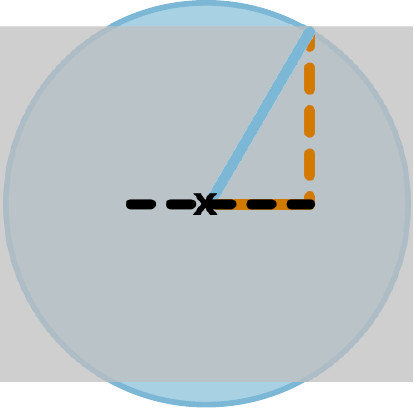}
    \put (-173, 15) { \tiny $t_{\max}$ }
    \put (-153.5, 24.5) { \tiny $\theta$ }
    \put (-168, 27) { \tiny $r$ }
    \put (-132, 20) { \footnotesize $r = \sqrt{ t^2_{\max} + \theta^2   }$}
    \put (-28, 15) { \tiny $t_{\max}$ }
    \put (-12, 28) { \tiny $\theta$ }
    \put (-24, 31) { \tiny $r$ }
    \put (-188.5, -4) { \tiny $t_{\max} \geq \theta$ }
    \put (-39, -4) { \tiny $t_{\max} < \theta$ }
    \caption{Definition of radius~$r$ (\textit{blue}) for the \ac{PC} clipping (step (i) of \ac{PCMA}) shown as side view with respect to the face (\textit{dashed black line}). 
    Radius~$r$ guarantees to enclose the entire face by enclosing the maximum distance $t_{\max}$ (\textit{dashed horizontal orange line}) of the \ac{COG} (\textit{black cross}) to face vertices. Besides, $r$ avoids to prefilter points by enclosing threshold~$\theta$ (\textit{dashed vertical orange line}) across the entire face. 
    The threshold band is marked in gray. 
    }
    \label{fig:sphere_radius}
\end{figure}

At first, we roughly reduce the search space for each face in order to accelerate the association.
To this end, we build a kD tree for the \ac{PC} (tile) and query the built tree with \acp{COG} of all faces.
Thereby, we detect all points within distance~$r$ for each face (ball query). 
The query parameter~$r$ is set in dependence of the manually set association threshold~$\theta$ and the maximum distance~$t_{\max}$ of the \ac{COG} to the respective face vertices.
Geometrically, $r$~is set to the length of the hypotenuse of the triangle as defined by $\theta$ and~$t_{\max}$. 
In simple terms, the query parameter~$r$ is set to the minimum distance that guarantees the manually set threshold~$\theta$ to be effective for the entire face while enclosing the entire face (cf.~\autoref{fig:sphere_radius}). 
Hence, $r$ prevents prefiltering of points by a too small spherical vicinity. 
The ball query delivers a subset of points, which may contain \textit{off-the-face points} and \textit{out-of-face points}. 

Second, we filter \textit{out-of-face points} by neglecting subset points whose orthogonal projections on the face plane are not enclosed by the face outline.
For details, we refer the interested reader to \cite{Laupheimer2020_ISPRS}.
Visually, the result of (ii) is the intersection of the spherical subset with radius~$r$ and the infinite triangular prism as defined by the face and its normal vector. We refer to this as the \textit{association prism}.

Finally, we filter the remaining \textit{off-the-face points}.
For this purpose, we calculate the orthogonal distance for each remaining point to the face plane. 
If the distance exceeds a chosen association threshold~$\theta$, the point is not associated with the face.
Since we have to compensate several discrepancies between \ac{PC} and mesh, we use a more sophisticated adaptive thresholding with an arbitrary user-defined number of filter levels~$n_l$. 

Each level~$l$ consists of two independent thresholds~$\theta_l^+$ and~$\theta_{l}^-$ limiting the association prism in the normal direction or the opposite direction respectively.
The absolute threshold values increase with ascending level.
Starting from level~1, the algorithm tries to associate points with the respective thresholds~$\theta_1^+$ and~$\theta_{1}^-$. 
If points have been linked, the association stops. 
Otherwise, the next level~$l$ is activated. 
This adaptive thresholding accelerates the association process. 
On the other hand, by nature of our approach, not all points might be associated.
Here, our reasoning is to favor near-surface points at the cost of missing to link a few points (fast small margin association).

The association information is stored as a per-point attribute.
For each associated point, the respective face index is attached to its attributes.
Non-associated points are marked with~$-1$.
The stored indices trivialize the transfer of features and labels from the mesh to the \ac{PC} (Mesh~$\mapsto$~PC in~\autoref{tab:association_options}).
In this case, we copy the desired values to the \ac{PC} at a stroke due to the one-to-many relationship. 
Reversely, the stored face indices can also be used to transfer features and labels from the \ac{PC} to the mesh (PC~$\mapsto$~Mesh in~\autoref{tab:association_options}).
However, to speed up the transfer, we directly couple the information transfer with the association mechanism.
The many-to-one relationship calls for information aggregation. 
For each face, we derive robust median features as gathered from the \ac{PC}. 
Features may embrace sensor-intrinsic and handcrafted features such as pulse characteristics and derived quantities \citep{Eitel2016}.
Analogously, majority votes determine the per-face labels as transferred from the associated points. 
Therefore, the association inherently is a label transfer tool and feature calculation tool (median features).
Non-associated faces are marked with~$-1$ and receive zeroed median features \citep{Laupheimer2020_ISPRS}.

\subsection{Image Mesh Association (ImgMA)}
\label{subsec:imgmeshassociation}
\begin{figure}[htbp]
    \centering    
    \begin{tabular}{cc}
         \multicolumn{2}{c}{
         \includegraphics[width=0.4\linewidth]{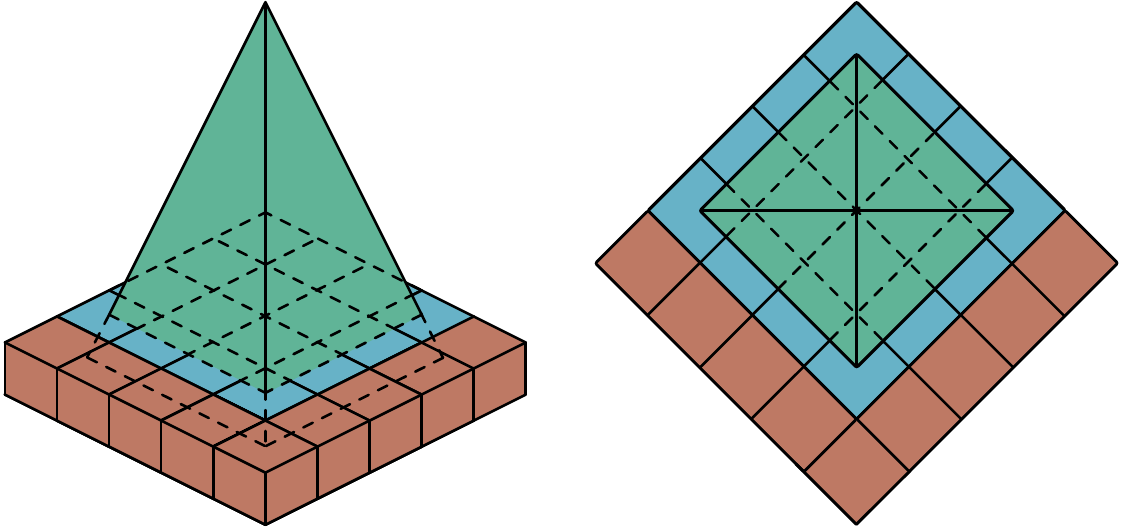} \vspace{-0.25cm}}
         \\
         \multicolumn{2}{c}{(I)}
         \vspace{0.25cm}
         \\
         \includegraphics[width=0.3\linewidth]{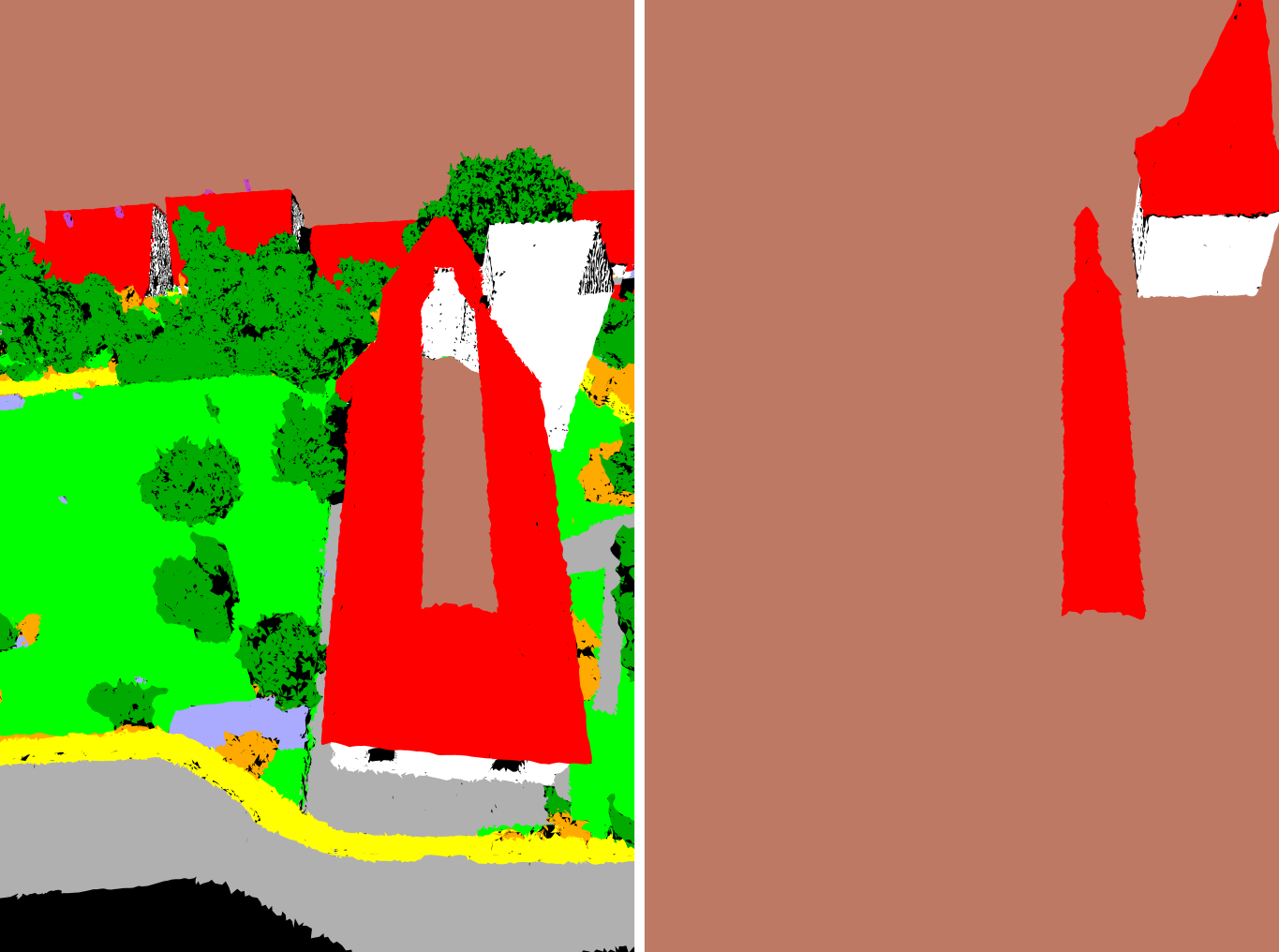} \hspace{0.5cm} & 
         \hspace{0.5cm} \includegraphics[width=0.3\linewidth]{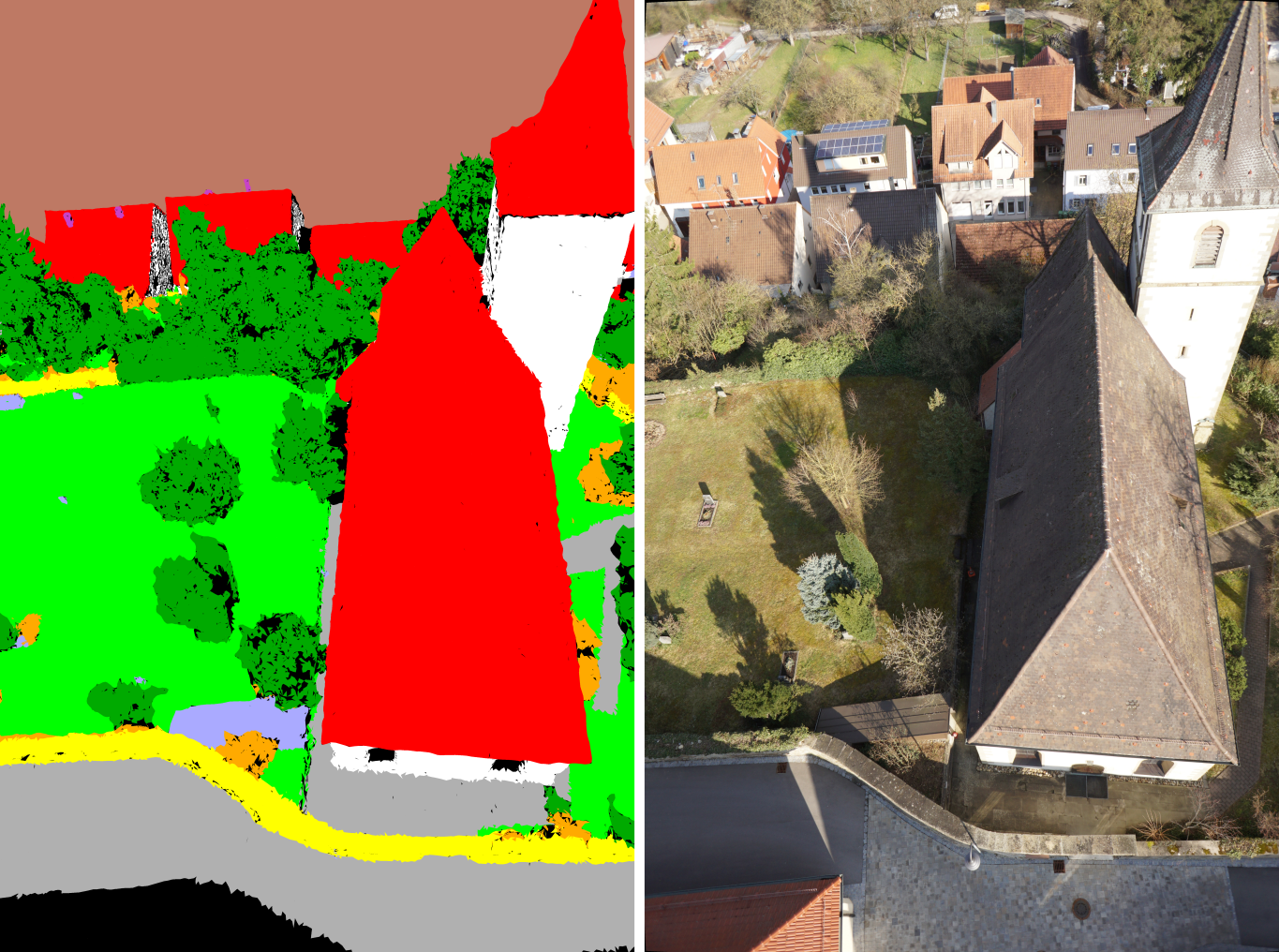}
         \vspace{-0.25cm}
         \\
         (II) \hspace{0.5cm} &  \hspace{0.5cm} (III)
    \end{tabular}
    \caption{
    Steps (I) - (III) of the \ac{ImgMA} shown in accordance with the information transfer Mesh~$\mapsto$~Img (cf.~\autoref{tab:association_options}). 
    The depicted example shows the association of an image (\textit{lower right}) with two vertically split mesh tiles of \ac{H3D}.
    \textit{(I)}: Preselection of visible mesh tiles per image shown schematically in isometric (\textit{left}) and top view (\textit{right}).
    The stretched camera pyramid (\textit{green}) intersects with some \acp{MBB} (\textit{blue)}. Non-intersecting \acp{MBB} are marked in \textit{reddish}. Dashed lines indicate non-visible parts.
    \textit{(II)}: Ray casting per image-tile-pair. 
    The \textit{reddish} area shows where ray casting fails due to missing intersections of image rays and considered mesh tiles. 
    \textit{Black} indicates intersection with unlabeled faces.
    \textit{(III)}: Final result after fusing ray casting results per image via depth filtering across tiles.
    }
    \label{fig:methodology_association_img_mesh}
\end{figure}

The \ac{ImgMA} explicitly links faces and pixels across various images in a geometry-driven approach. 
To accelerate the \ac{ImgMA} we make use of the given mesh tiling. 
Each pixel is assigned with the visible face as detected by the following three steps:
(I) preselection of visible mesh tiles per image utilizing \acp{MBB} of the tiles,
(II) ray casting per image and tile (image-tile-pair), 
and (III) fusion of ray casting results per image via depth filtering across tiles.
We end up with $n_\mathit{px}$ associated pixels across $n_\mathit{img}$ images for each face. 
\autoref{fig:methodology_association_img_mesh} shows the workflow by means of an oblique example image and two vertically separated tiles.

For each image, we first detect visible tiles and perform the subsequent ray casting procedure in a parallelized fashion for the subset of tiles only.
We define a tile to be visible if its \ac{MBB} intersects with the stretched camera pyramid (\ac{MBB} visibility check).
The stretched camera pyramid is defined by the projection center and the projection rays crossing the corner pixels of the respective image. 
The lowest of all \ac{MBB} faces limits the stretched camera pyramid. 
Since the \acp{MBB} are not fully occupied by the enclosed mesh tiles, some detected tiles might not contain any visible faces.
Nonetheless, this approach significantly reduces the number of tiles that have to be processed by the ray casting procedure for each image.
We detect intersections of the camera pyramid and a \ac{MBB} by a three-stage check starting with the most likely and fastest test.
At first, we check for each tile if any corner point of the respective \ac{MBB} is inside the camera pyramid (point in polyhedron test). 
The second and third stage perform edge face intersections, checking intersections of pyramid edges starting from the projection center and any \ac{MBB} face or intersections of \ac{MBB} edges and any pyramid face.
Once a test succeeds, the respective enclosed tile is marked as visible and the residual checks are omitted (check omission). 

As a result of (I), we receive a list of visible tiles for each image, i.e. each image~$\mathit{img}$ is linked to $n_\mathit{tiles}^\mathit{img}$~tiles.
Hence, there are $n_\mathit{tiles}^\mathit{img}$~image-tile-pairs for each~$\mathit{img}$.
Vice versa, a list of visible images for each tile is stored. 
For each image-tile-pair, visible faces are determined via ray casting. 
For this purpose, for each pixel, a 3D projection ray is created and intersected with the mesh faces of all linked tiles. 
The intersected faces are candidates for the final association result.
Concerning a single image-tile-pair, all candidate faces are truly visible. 
However, an image probably covers multiple tiles, and consequently, some faces might be occluded by faces of another tile (cf.~\autoref{fig:methodology_association_img_mesh}).
For the final result, we fuse the $n_\mathit{tiles}^\mathit{img}$ ray casting results across the visible tiles into one final ray casting result per image.

Implementationally, the fusion across the $n_\mathit{tiles}^\mathit{img}$ image-tile-pairs is done implicitly by depth updates. 
Initially, each pixel is associated with a near-infinite depth value. 
The association information is updated whenever a candidate face reduces the depth value. 
Hence, the final association information is steered by faces of minimum depth that are truly visible (i.e. faces that mark the first intersection along the respective ray). 
To speed up the implementation, steps (II) and (III) are parallelized with respect to images.
Each process handles $n_\mathit{tiles}^\mathit{img}$~image-tile-pairs per image.
The implicit fusion reduces the memory footprint of the algorithm since only the final ray casting result has to be stored. 

To minimize the memory footprint, we avoid storing the association information channel-wise due to the curse of dimensionality. 
Particularly for oblique images, only a small part of the image may be associated due to the limited reconstruction area of the mesh. 
Instead, we store the final association information as a \textit{sparse pixel cloud} per image consisting only of pixels that have been linked with a face.
The sparse pixel cloud contains tuples of associated pixel positions, the depth, the tile-dependent face index, and optionally, other attributes (e.g. labels) as transferred from the mesh.

The stored face indices trivialize the transfer of features and labels from the mesh to the images (Mesh~$\mapsto$~Img in~\autoref{tab:association_options}). 
We copy the desired quantities to the linked pixels of the associated images at a stroke due to the one-to-many-to-many relationship. 
Reversely, the many-to-many-to-one relationship calls for information aggregation (Img~$\mapsto$~Mesh in~\autoref{tab:association_options}). 
For each face, we derive robust median features as gathered from the associated pixels across the respective images. 
Features may embrace sensor-intrinsic multi-spectral information, handcrafted features, and features as derived by \ac{DL} pipelines.
Analogously, majority votes determine the per-face labels as transferred from the linked pixels.

\subsection{Point Cloud Image Association (PCImgA)}
\label{subsec:pcimgassociation}
The \ac{PCImgA} aims for the linking of pixel locations and 3D points. 
Theoretically, the collinearity equations establish an explicit relationship between 3D points and pixels. 
Each point can be projected into the image space given the exterior and interior orientation of the respective image. 
However, the bare projection cannot check for visibility, and hence, links visible and non-visible points with imagery. 
Therefore, point visibility has to be checked prior to the linking of pixels and 3D points. 
To this end, we leverage the mesh representation by combining mechanisms \ac{PCMA} and \ac{ImgMA} implicitly and explicitly.
The implicit linking is face-centered whereas the explicit linking is point-centered or pixel-centered (dependent on the transfer direction).
Making a detour via the mesh largely solves the visibility problem for 3D points in image space.

The implicit linking couples the association mechanisms \ac{PCMA} and \ac{ImgMA} by simply executing them sequentially. 
Specifically, $n_\mathit{px}$~pixels of $n_\mathit{img}$~images and $n_\mathit{pts}$~points are exclusively linked to the respective face. 
For the information transfer, the information from the starting representation is gathered per face and transduced to the target representation. 
Therefore, the joint face apparently establishes a linking of points and pixels.
The per-face label and features as derived from the starting representation are determined via majority voting and median aggregation respectively (cf.~\autoref{tab:association_options}). 
Subsequentially, the per-face aggregations are copied to the target modality (one-to-many relationship).

On the contrary, the explicit linking couples the association mechanisms \ac{PCMA} and \ac{ImgMA} by leveraging the stored association information and utilizing the collinearity equations. 
As a result of \ac{ImgMA}, visible faces for each image are known. 
At the same time, mechanism \ac{PCMA} delivers the associated points for each face. 
Consequently, associated points of visible faces are marked as visible.
Therefore, the combination of \ac{PCMA} and \ac{ImgMA} results in a visible subset of the \ac{PC} per image.
The collinearity equations explicitly link the visible points and the pixel locations across all imagery.
Therefore, explicit linking truly associates points and pixel locations. 

For each point, there is an unambiguous pixel location in each image. 
However, the reverse situation is ambiguous.
Depending on \ac{GSD} and point density, each pixel of each image may enclose several visible points. 
Therefore, transferring information from the \ac{PC} to imagery (\ac{PC}~$\mapsto$~Img in \autoref{tab:association_options}) demands a pixel-wise aggregation (many-to-one relationship per pixel and image). 
The features and labels are aggregated by median aggregation and majority voting respectively.
To accelerate the process, we approximate the pixel-wise aggregation by transferring only information of the point of minimum depth. 
This approximation simplifies the association to a one-to-one relationship per pixel and image.
Likewise, a one-to-one relationship holds if the \ac{GSD} is smaller than the point distance.
Since each point is covered by several images, the information transfer Img~$\mapsto$~\ac{PC} requires a point-wise aggregation across images.

\subsection{Preconditions and Limitations}
\label{subsec:preconditions_limitations}
The proposed method connects 3D \acp{PC}, photographic imagery (following the central perspective), and textured meshes. 
By nature of the algorithm, it merely depends on the pure existence of those three modalities. 
Therefore, it is a generic approach that works with any photographic image and \ac{PC} regardless of the acquisition platform (aerial, terrestrial, mobile), the image type (panchromatic, RGB, multi-spectral), and \ac{PC} type (\ac{MVS} cloud, LiDAR cloud, persistent scatterer cloud).
However, we focus on the linking of aerial RGB imagery and \ac{ALS} \acp{PC} along with the respective textured 3D mesh (cf.~\autoref{sec:results_discussion}). 
Furthermore, photogrammetric meshes, LiDAR meshes, or hybrid meshes can be processed.
The linking is not constraint to a specific mesh generation algorithm or mesh geometry (2.5D or 3D). 
However, the entire association benefits from good geometric reconstruction.
We are aware of the fact that proper meshing of our complex world is a hard task and still subject to research.

A proper reconstruction ensures an appropriate entity linking and information transfer. 
As a general rule, the better the mesh represents the true 3D structure of the real world, the better works the proposed association mechanism and the subsequent information transfer.
Obviously, as a precondition to the association, the considered data representations have to cover the same area and have to be oriented in the same coordinate system. 

\begin{figure}[htbp]
    \centering
    \begin{tabular}{c c}
        \multicolumn{2}{c}{\includegraphics[width=0.48\columnwidth]{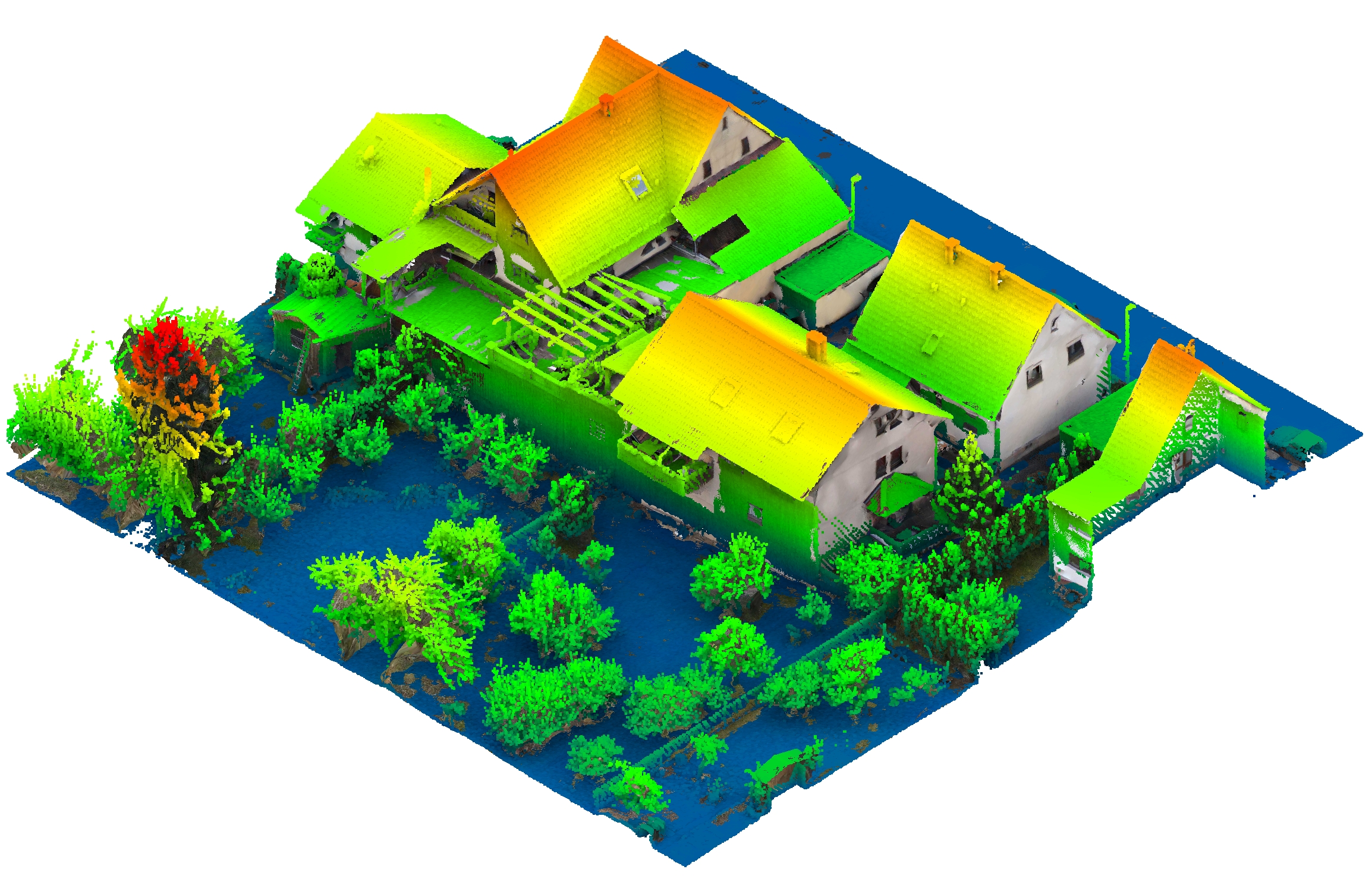}} \\
        \includegraphics[width=0.28\columnwidth]{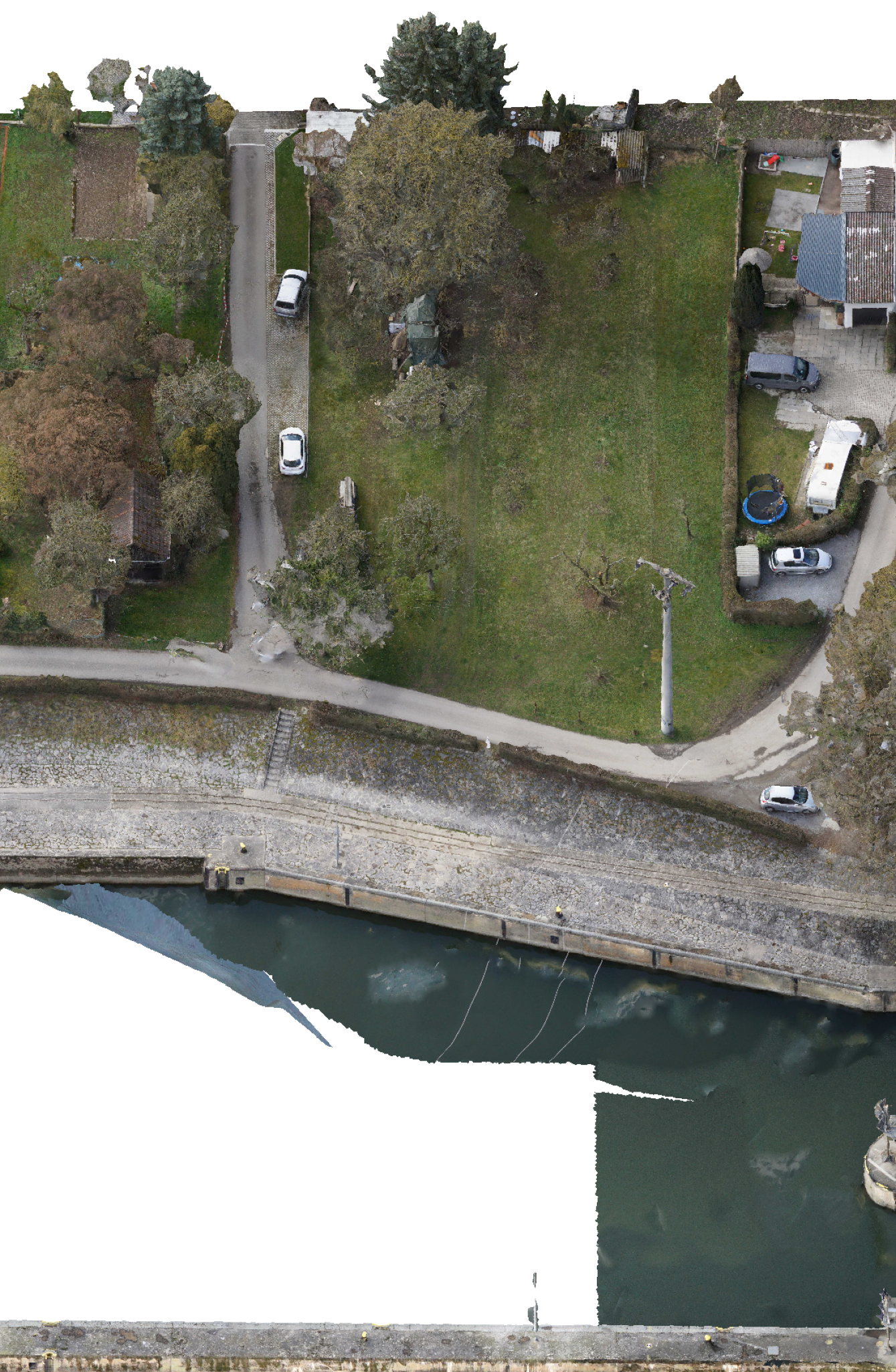} 
        \hspace{0.5cm}
        &
        \hspace{0.5cm}
        \includegraphics[width=0.28\columnwidth]{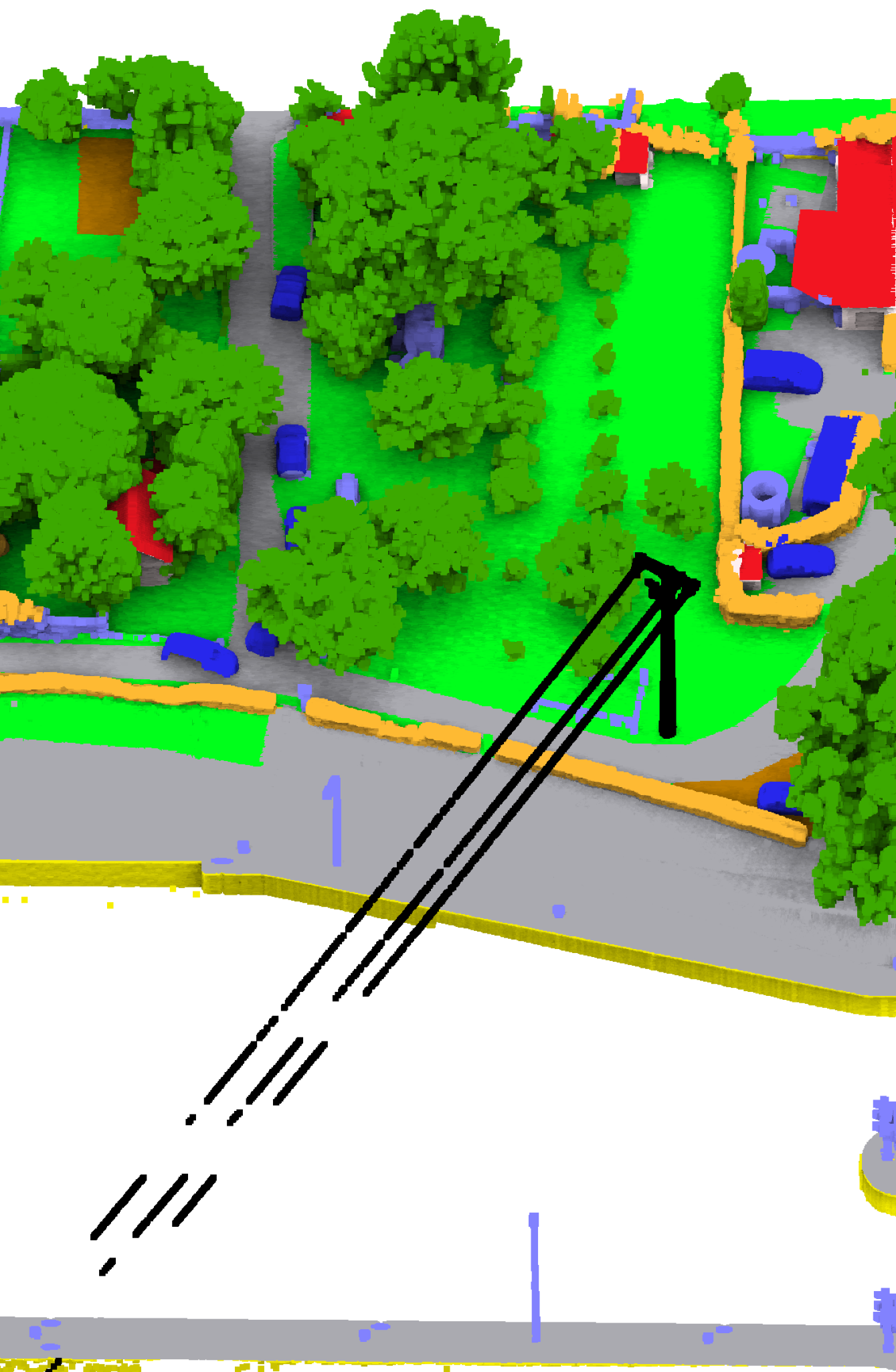} 
    \end{tabular}
    \caption{Discrepancies and data gaps in 3D models of \ac{H3D}. 
    \textit{Top}: Urban area. Overlay of textured mesh and \ac{ALS} \ac{PC} (height-coded). 
    Facades are reconstructed entirely in the mesh but are captured only partially by the \ac{PC}. 
    \textit{Bottom}: Ship lock area. The mesh (\textit{left}) partially reconstructs the river but misses to fully reconstruct thin structures like light poles and power lines, which are captured entirely in the \ac{PC} (\textit{right}). The asynchronous data acquisition of images and \ac{ALS} data cause inconsistencies between mesh and \ac{PC} (e.g. for cars).
    }
    \label{fig:discrepancies_interrepresentation}
\end{figure}

Assuming a proper co-registration, entities across representations can only be associated when underlying real-world objects are captured or reconstructed in all representations. 
In other words, each face should at least enclose one point or one pixel respectively. 
Reversely, each point or pixel should be mapped onto a corresponding face.
However, these relationships do not always exist due to inter-representation differences such as object penetration (cf.~\autoref{sec:intro}, \citeauthor{Laupheimer2020_ISPRS}, \citeyear{Laupheimer2020_ISPRS}).
For instance, due to occlusion during data acquisition, a facade may not be captured in the \ac{ALS} data, and hence the respective faces cannot link any points.
Reversely, due to mesh simplification, a thin-structured object may not have been reconstructed entirely in the mesh, but the object is fully captured in the \ac{PC} and imagery.
Additionally, there might be inter-representation discrepancies due to asynchronous data acquisition.
\autoref{fig:discrepancies_interrepresentation} shows both situations for the \ac{H3D} data set.

\begin{figure}[htbp]
    \centering
    \includegraphics[width=0.75\columnwidth]{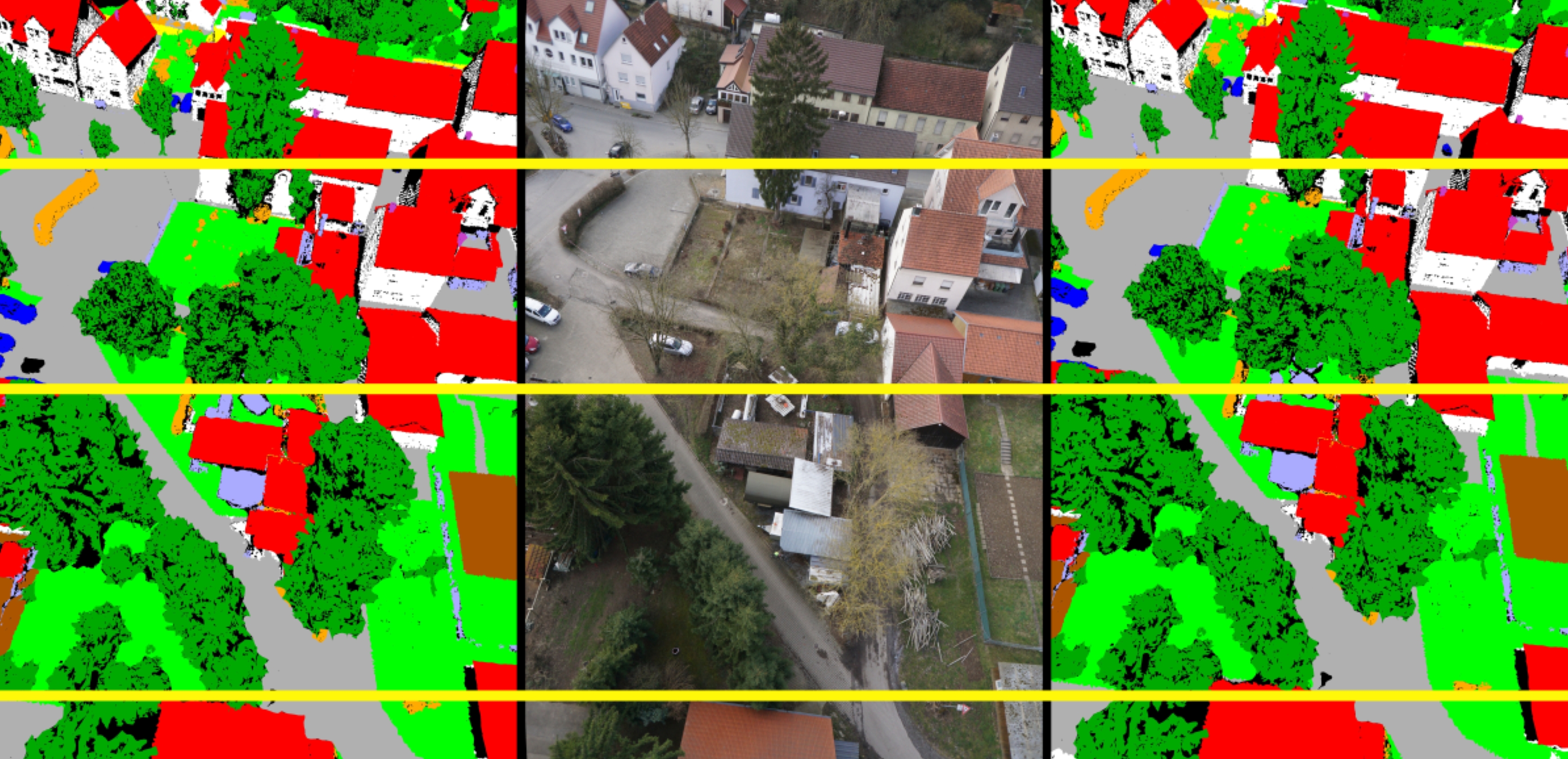}
    \caption{Influence of the co-registration quality as visualized by the label transfer (\textit{left/right}: good/bad relative orientation of 3D data and imagery). The yellow lines depict "epipolar lines" crossing roof corners in the RGB image (\textit{center}). The orientation parameters as achieved by bundle adjustment have been slightly falsified artificially for the visualization on the right.}  
    \label{fig:img_labels}
\end{figure}

Misalignment among modalities is a decisive issue for multi-modality, wherefore proper co-registration is subject to current research. 
Ideally, imagery and \ac{PC} data are co-registered simultaneously in a joint adjustment. 
Consequently, the derived mesh is aligned with both data sources and co-registration issues are obsolete.
Nonetheless, reality shows that co-registration discrepancies are an important and real issue (cf.~\autoref{sec:data}). 
A good relative orientation of 3D data and imagery is beneficial to the linking of pixels with points and faces. 
We are aware of the fact that \ac{PCImgA} and \ac{ImgMA} depend on the quality of the co-registration. 
However, the proper co-registration of both data sources is not the focus of this work. 
\autoref{fig:img_labels} depicts the influence of the co-registration quality of imagery and 3D data.

Inherently, \ac{MVS} meshes and \ac{MVS} \acp{PC} are perfectly aligned with imagery. 
Hence, incorrect or missing associations between the representations are only due to the reconstruction quality and data gaps (as for~\ac{V3D}). 

\begin{figure}[htbp]
    \centering
    \includegraphics[width=0.25\columnwidth]{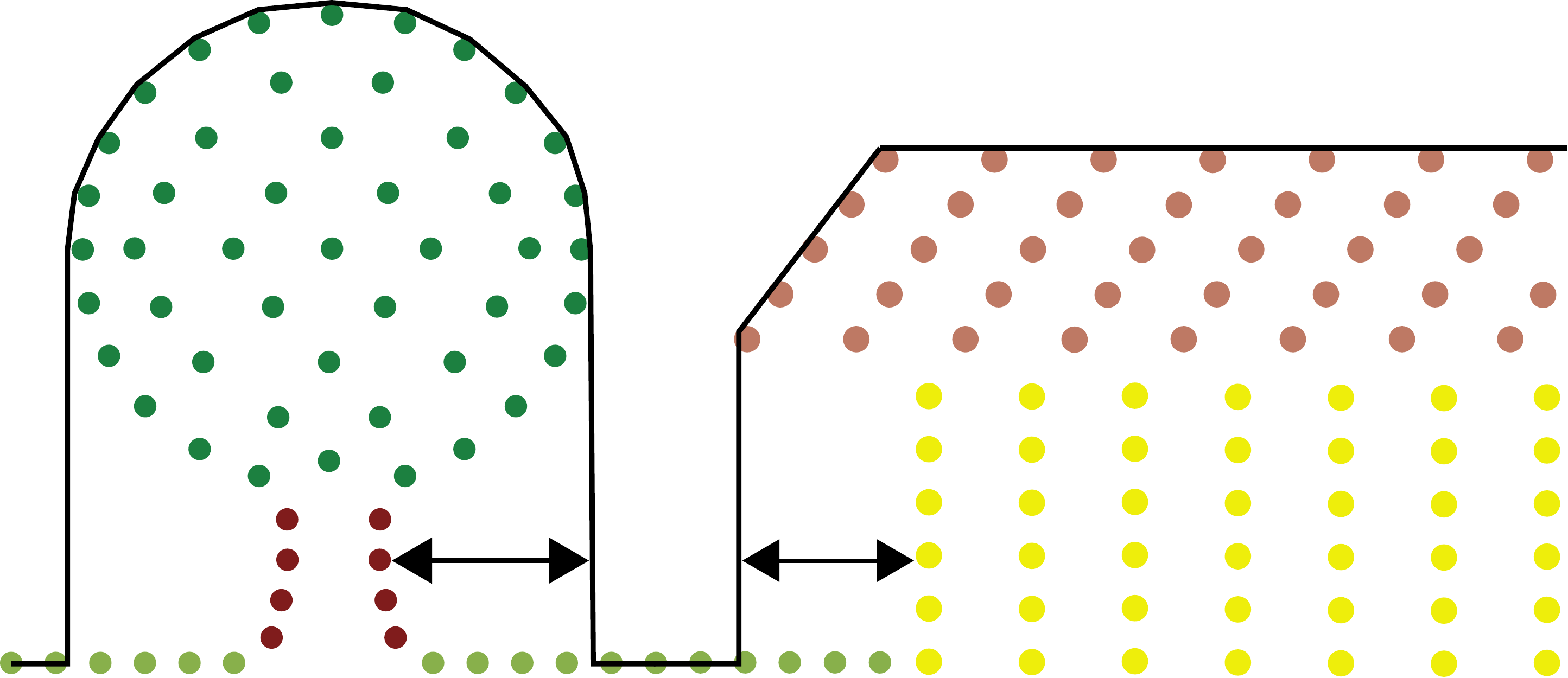}
    \hspace{0.5cm}
    \includegraphics[width=0.25\columnwidth]{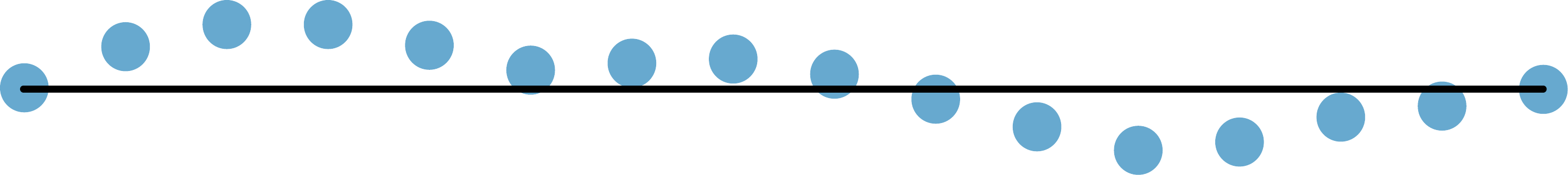}
    \hspace{0.5cm}
    \includegraphics[width=0.25\columnwidth]{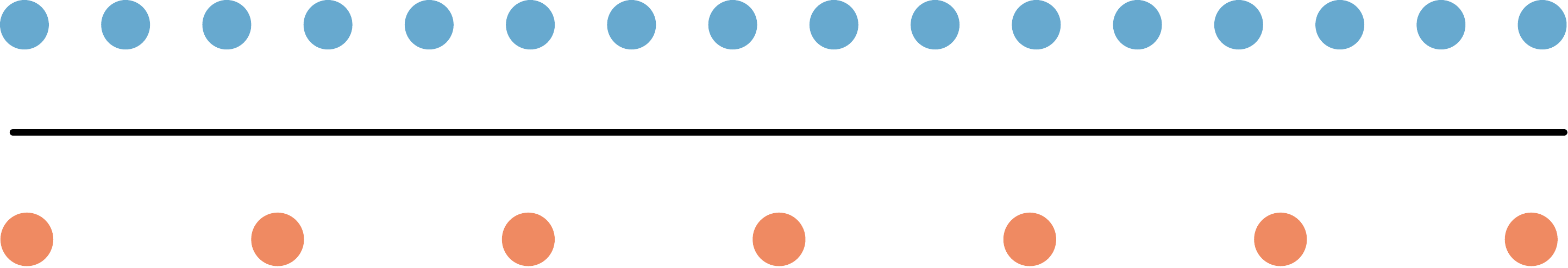}
    \caption{Discrepancies between \acp{PC} and the mesh (\textit{black line}) caught by adaptive thresholding.
    \textit{Left}: \textit{Black arrows} indicate the discrepancy between the 2.5D mesh and the annotated 3D \ac{PC}. 
    \textit{Center}: The noisy \ac{PC} (\textit{blue}) oscillates about the reconstructed mesh.
    \textit{Right}: There might be misalignment between \acp{PC} (\textit{blue}: \ac{MVS}, \textit{orange}: LiDAR) and the mesh as generated of a single source or complementary sources.}
    \label{fig:issue_pcmeshassociation_schematic}
\end{figure}

Since the relationship of 3D space and image space is strictly defined by the collinearity equations, we highlight the discussion of \ac{PCMA}. 
\autoref{fig:issue_pcmeshassociation_schematic} and \autoref{tab:points_unassociated} sketch discrepancies of \ac{PC} and mesh despite representing the same real-world objects. 
These discrepancies are largely covered by the adaptive thresholding.
Despite and due to this technique, not all points are associated with faces. 
There are three groups of unassociated points (cf.~\autoref{tab:points_unassociated}).

\begin{table}[htbp]
    \resizebox{\textwidth}{!}{
    \centering
    \footnotesize
    \begin{tabular}{c c}
        \toprule
         \makebox[8cm][l]{\normalsize A) points outside the threshold range} &  \\
         \midrule
         \includegraphics[width=0.5\columnwidth]{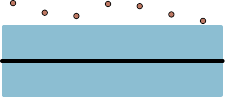}
          & 
         \includegraphics[width=0.5\columnwidth]{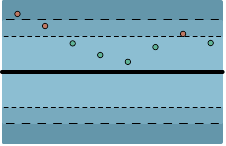} \\
         A1: points outside the association prism &
         A2: points in different threshold bands ("early stopping") \\
         \midrule
         \makebox[8cm][l]{\normalsize B) points outside the association prisms} & \\
         \midrule
         \includegraphics[width=0.35\columnwidth]{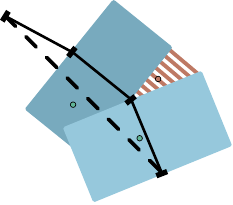}
          & 
         \includegraphics[width=0.5\columnwidth]{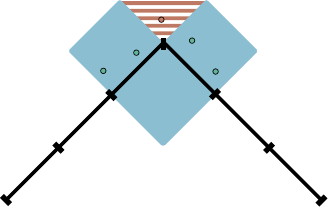} \\
         B1: noisy measurements and noisy reconstruction &
         B2: misalignment \\
         \midrule
         \makebox[8cm][l]{\normalsize C) points on the association prism (optional)} & \\ 
         \midrule
         \includegraphics[width=0.25\columnwidth]{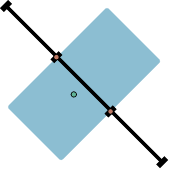}
          & 
         \includegraphics[width=0.25\columnwidth]{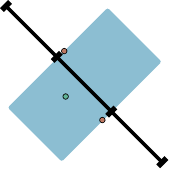}   \\
         C1: points coincide with face vertices &
         C2: points coincide with association prism boundary\\
         \bottomrule
    \end{tabular}
    } 
    \caption{The schematic drawings illustrate the three cases where faces (\textit{black} lines, separated by \textit{black} strokes) and points are not linked (side view with respect to the face). 
    Non-associated points are marked in \textit{red}, associated points in \textit{green}. 
    The association prisms are marked in \textit{blue}. 
    Adaptive thresholds are omitted - except for~A2.
    A2~depicts the increasing thresholds with increasing blueness. 
    Diverging association prisms create \textit{dead zones} like depicted in~B (\textit{hatched in red}). 
    B1~mimics a perfect planar surface as \textit{dashed black} line.}
    \label{tab:points_unassociated}
\end{table}

Visually, the association is done utilizing the per-face association prism which is limited by the range of manually set thresholds~$\theta_l^+$ and~$\theta_{l}^-$. 
The threshold-based approach filters all points whose orthogonal projections to the face plane exceed a specific value (cf.~\autoref{tab:points_unassociated}, A1). 
This prevents the linking of points and faces that most likely represent different surfaces.
Furthermore, the adaptive thresholding breaks the association once a point is associated at any threshold level~$l$ ("early stopping"). 
Hence, points that are closer to the face than the maximum threshold may not be linked (cf.~\autoref{tab:points_unassociated}, A2).
Levels~$l>1$ can be seen as a fall-back for scenarios where a proper association is not possible.
The adaptive thresholding favors near-surface points by ensuring the association of points fulfilling the smallest threshold.

The adaptive thresholding facilitates a varying degree of freedom by the set thresholds and their inter-level spacing. 
Eventually, it balances the strictness of the point-face linking.
Small-valued thresholds along with small-spaced levels enforce a tight coupling where only points close to the mesh surface are associated. 
Large values along with large level spacing loosen the coupling. 
Moreover, the asymmetric two-fold thresholding per level enables non-symmetric filtering improving flexibility and adaptiveness. 
Therefore, the presented association mechanism is agnostic to the geometric structure of mesh geometry: 2.5D or 3D meshes can be processed. 
The asymmetric adaptive thresholding allows associating faces with points where 2.5D and 3D geometry differ significantly while favoring the association of near-surface points (e.g. facades or tree stems, cf.~\autoref{fig:issue_pcmeshassociation_schematic} on the left).
\cite{Laupheimer2020_ISPRS} discuss in detail the particular challenges for the \ac{PCMA} using a 2.5D mesh. 

The set of association prisms does not enclose all points of the \ac{PC}.
Particularly, points above and below the mesh surface may fall into \textit{dead zones} not covered by the association prisms. 
The prisms of adjacent faces diverge when they form convex or concave surfaces, i.e. their normal vectors are not parallel. 
Non-perfect reconstructions of planar surfaces artificially introduce convex or concave structures (cf.~\autoref{tab:points_unassociated}, B1).
Naturally, points above truly convex surfaces or below concave surfaces cannot be linked. 
Typically, points above the reconstructed roof ridge cannot be associated (cf.~\autoref{tab:points_unassociated}, B2).
For this reason, co-registration discrepancies increase the number of non-associations.

The previously described missed associations are owed to the nature of the problem itself and the implementation aiming for a good trade-off of memory and speed (by adaptive thresholding). 
Besides, we declare points that are projected on the face edges or vertices to be \textit{out-of-face points} (cf.~\autoref{tab:points_unassociated}, C). 
By these means, we avoid their linking by choice.
Technically, these points belong to two adjacent faces~A and~B.
Hence, it is hard to decide whether to assign them to face~A or its adjacent face~B. 
Therefore, such points cannot be linked unambiguously and may cause ambiguity in the information transfer. 
Here, our reasoning is to link not all, but unambiguous points. 
As a side-effect, neglecting these points accelerates the association process.
However, if desired, our implementation allows us to link these points, too \citep{Laupheimer2020_ISPRS}.

To the best of our knowledge, meshing algorithms depend fully on geometry and do not incorporate semantics. 
Therefore, reconstructed faces do not necessarily represent semantic borders.
For instance, consider the transition of a planar \textit{impervious surface} to \textit{green space} in the real world. 
The mesh representation may simplify this scenario to a single large face. 
On the contrary, the same scenario is captured properly in the \ac{PC} and imagery.
Consequently, too large triangles at class borders will associate points and pixels of different classes (cf.~\autoref{fig:inconsistency_overview}). 
For this reason, semantically incorrect associations are unavoidable due to the meshing. 

\begin{figure}[htbp]
    \centering
    \begin{tabular}{c c}
        \includegraphics[width=0.425\columnwidth]{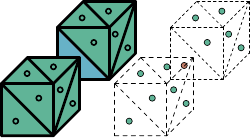} \hspace{0.7cm} &
        \hspace{0.7cm} \includegraphics[width=0.2\columnwidth]{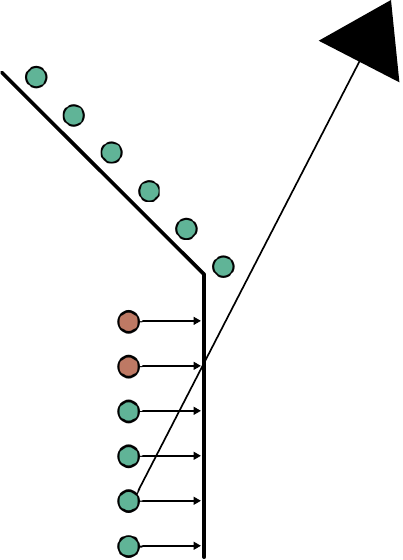} \\
         (a) \hspace{0.7cm} &   \hspace{0.7cm} (b)
        \put (-245.5, 13.5) {\small opaque}
        \put (-186.5, 13.5) {\small transparent}
    \end{tabular}
    \caption{ 
    The sketches indicate truly (\textit{green}) and apparently visible points (\textit{red}) as detected by \ac{PCImgA} utilizing the mesh (\textit{black wireframe}) as a proxy. 
    (a) showcases an apparently visible point due to being linked to a truly but non-fully visible face (\textit{blue}).  
    Fully visible faces are depicted in \textit{green}. 
    The dashed cubes on the right mimic the situation of the opaque cubes in "transparent" mode.
    (b) showcases apparently visible points due to the divergence of normal direction and line-of-sight.
    Points are associated along the normal direction of faces. Hence, visible faces might be linked to truly non-visible points. 
    }
    \label{fig:association_pc_img_visibility}
\end{figure}

Naturally, the \ac{PCImgA} inherits the limitations of its sub pipelines. 
Besides, there are specific issues regarding the \ac{PC} visibility as derived via the mesh as a proxy. 
Faces are marked as visible once a pixel is associated with the face. 
Points, in turn, are marked as visible if they are associated with a visible face. 
However, a face marked as visible does not have to be entirely visible. 
Moreover, a face marked as visible does not have to be associated with points that are truly visible as well. 
The adaptive thresholding links points and faces along the normal directions whereas the line-of-sight is relevant for the point visibility. 
In other words, taken individually, the made associations by the sub pipelines are correct, but their composition does not guarantee a correct visibility check for each point-pixel relationship (using explicit \ac{PCImgA}).
On the contrary, the implicit \ac{PCImgA} makes use of the correct point-face and face-pixel linking. 
Rephrased, the implicit linking overcomes georeference issues utilizing the adaptive thresholding and uses the correct visibility checks for faces.
\autoref{fig:association_pc_img_visibility} depicts point visibility and showcases apparently visible points. 
It is unlikely that all truly non-visible points that are marked as visible are occluded by truly visible points. 
For this reason, depth filtering helps only when truly non-visible and truly visible points are on the same projection ray. 
Otherwise, truly non-visible points are linked with pixels through collinearity equations causing incorrect information transfer (for the explicit linking of the \ac{PCImgA}). 
Generally, smaller faces better represent the geometry and reduce the impact of truly non-visible points.
However, large triangles are beneficial concerning processing time.

\section{Results and Analysis}
\label{sec:results_discussion}
\begin{figure}[htbp]
    \centering
    \includegraphics[width=0.95\linewidth]{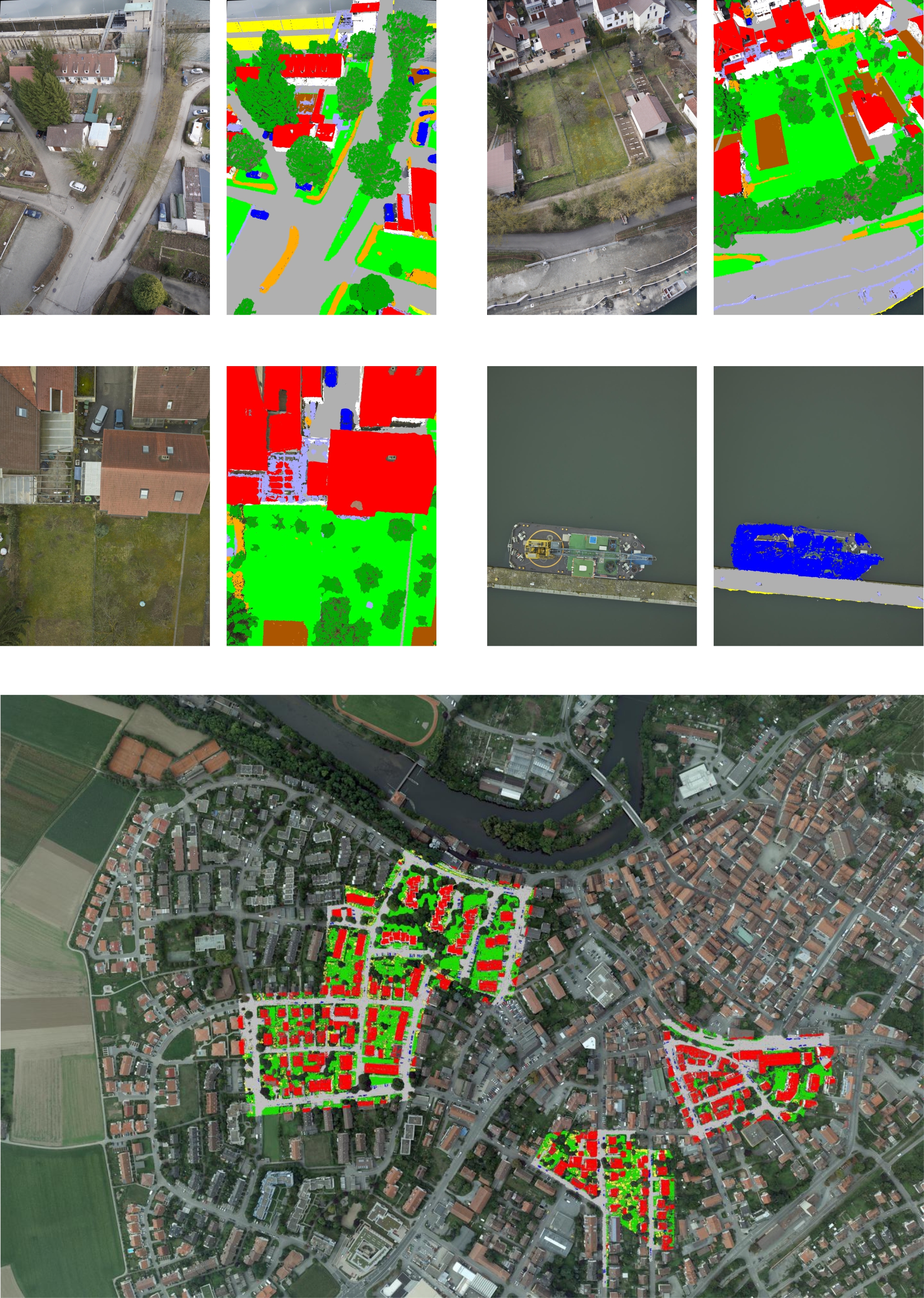}
    \caption{
   Automatically annotated images with labels as transferred from the \ac{PC} to the mesh and, therefrom, to image space (implicit \ac{PCImgA}) for \ac{H3D} (\textit{top}: oblique, \textit{center}: nadir) and \ac{V3D} (\textit{bottom}). 
   Non-associated pixels are depicted with RGB values. Original RGB images are shown on the left except for the nadir image of \ac{V3D} which covers the entire labeled area.
   }
    \label{fig:results_imagery}
\end{figure}

We demonstrate the capability of our methodology, its flexibility, and adaptiveness to underlying data by deploying \ac{V3D} and \ac{H3D} (cf.~\autoref{sec:data}).
To quantitatively analyze the linking methodology, \ac{GT} data is necessary for each modality. 
However, to the best of our knowledge, there is no real-world data set that provides annotations for \ac{PC}, mesh, and imagery at the same time (cf.~\autoref{subsec:relatedwork_gt}). 
For this reason, we qualitatively verify the effectiveness of the explicit entity linking visualizing the label transfer.
We visualize the achieved explicit entity linking by transferring labels from the manually annotated \acp{PC} to the mesh, and therefrom, to image space.
However, we want to emphasize that the transferred information is not limited to labels. 
Features can be transferred to other modalities, too.
\autoref{fig:juggler_alternative} exemplarily shows the annotated modalities for a dedicated tile from \ac{H3D} as achieved by the proposed methodology. 
\autoref{fig:results_imagery} shows a selection of automatically annotated images of various \acp{GSD} as derived via the implicit \ac{PCImgA}. 
We opt for the implicit linking to create densely labeled images since faces are projected to image space instead of single points. 
Furthermore, for \ac{V3D}, implicit \ac{PCImgA} makes use of the perfect co-registration of the \ac{MVS} mesh and imagery.
At the same time, the enclosed \ac{PCMA} is able to dampen the co-registration discrepancy in 3D space by leveraging the adaptive thresholding. 
Hence, the implicit linking avoids inconsistent point-pixel-pairs.
Both figures visually verify that the transfer operates reasonably and smoothly on both data sets featuring different scales and resolutions. 
Besides, \autoref{fig:results_imagery} reveals the dependence of synchronous acquisition and mesh quality.
The picture on the center-left shows different positions of a car due to asynchronous capturing of nadir imagery and \ac{ALS} data (cf.~\autoref{sec:data}).
The picture on the center-right depicts a ship (class \textit{vehicle}) in the ship lock surrounded by water. 
Since the mesh does not properly reconstruct the ship, the transferred labels do not cover the entire ship in image space. 
   
Despite the absence of \ac{GT} for all modalities, we try to quantify the entity linking by a proxy analysis: forward and backward passing of labels.
The relationship of image space and 3D space is well-known and strictly defined by collinearity equations and hence, does not have to be validated.
For this reason, we highlight the label transfer from the \ac{PC} to the mesh (forward pass) and therefrom back to the \ac{PC} (backward pass). 
During forward passing, we aggregate labels on the face-level via majority vote. 
The backward pass is a straightforward copy operation utilizing the stored association information. 
The comparison of back-transferred and manual annotations allows us to validate the effectiveness of~\ac{PCMA} (label consistency check).    

For the used data sets, 
we found in an empirical process the association to perform best with thresholds~$\theta_1=\pm\SI{30}{cm}$, $\theta_2=\pm\SI{60}{cm}$, $\theta_3=\pm\SI{120}{cm}$ (\ac{V3D}) and~$\theta_1=\pm\SI{5}{cm}$, $\theta_2=\pm\SI{10}{cm}$, $\theta_3=\pm\SI{15}{cm}$~(\ac{H3D}).
The chosen thresholds are guided by the shift between mesh and \ac{ALS} data (cf.~\autoref{sec:data}). 
In particular, thresholds are fine-tuned to maximize the number of associations while keeping mismatches of back-transferred and manual labels at a minimum.

\begin{figure}[htbp]
    \centering
    \begin{tabular}{cc}
        \includegraphics[width=0.45\columnwidth]{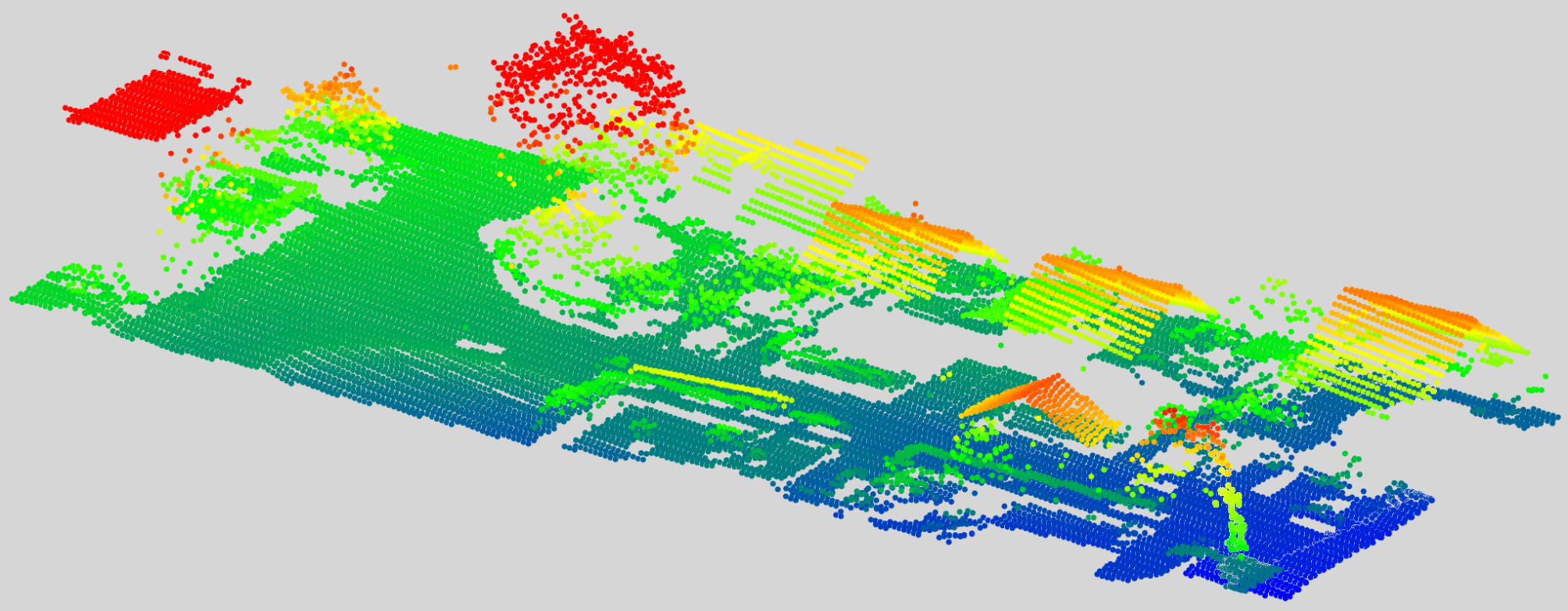} 
        \hspace{-0.215cm}  & \hspace{-0.215cm} 
        \includegraphics[width=0.45\columnwidth]{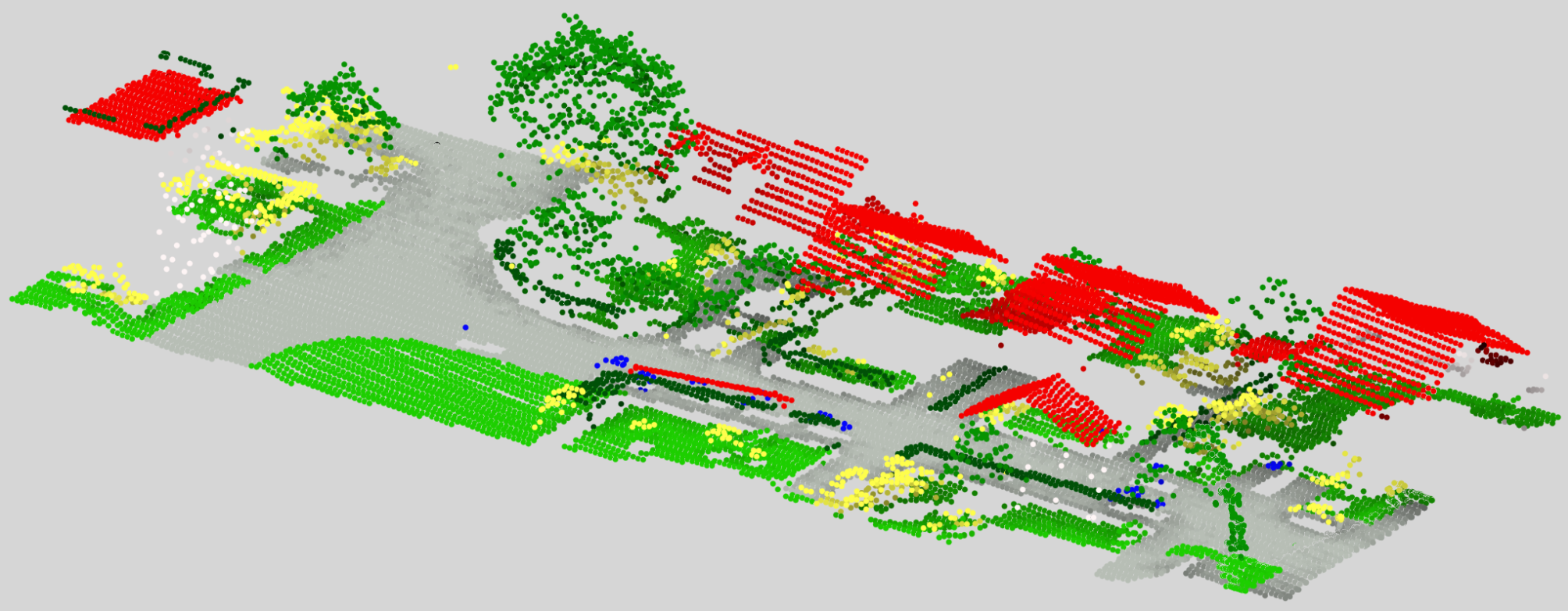}
        \\
        \includegraphics[width=0.45\columnwidth]{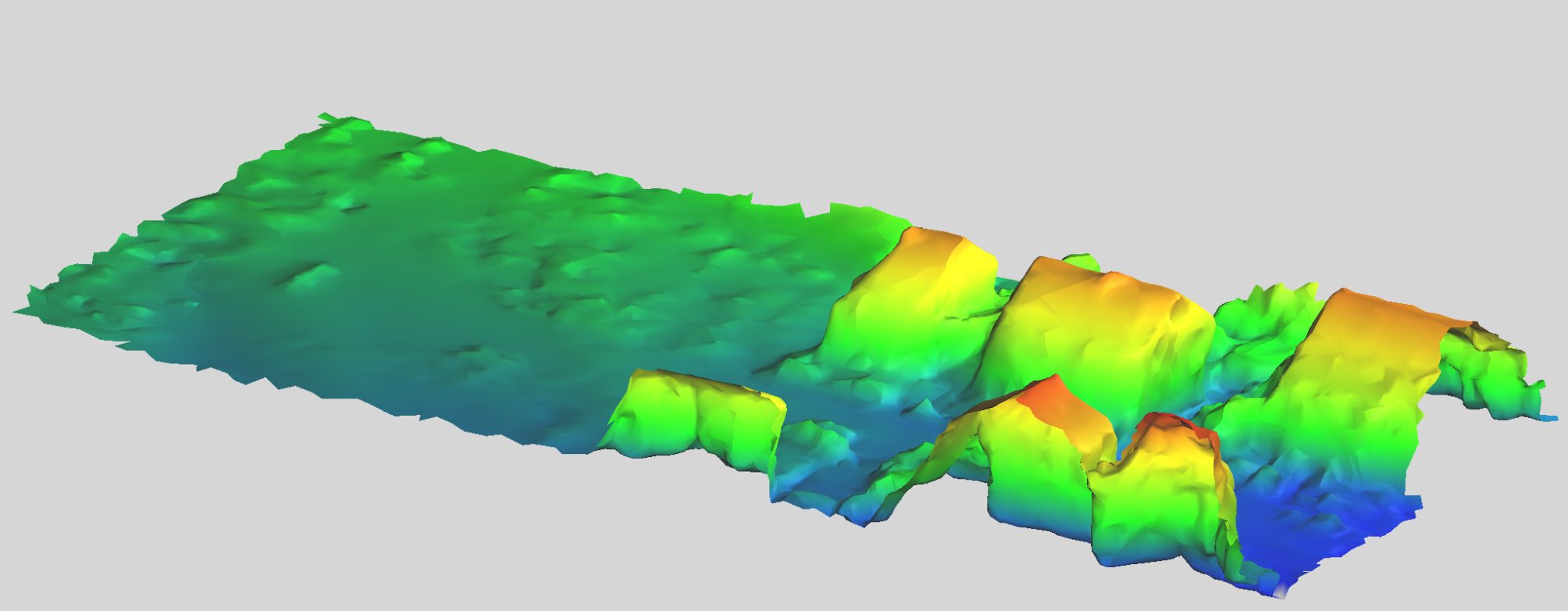}
        \hspace{-0.215cm}  & \hspace{-0.215cm} 
        \includegraphics[width=0.45\columnwidth]{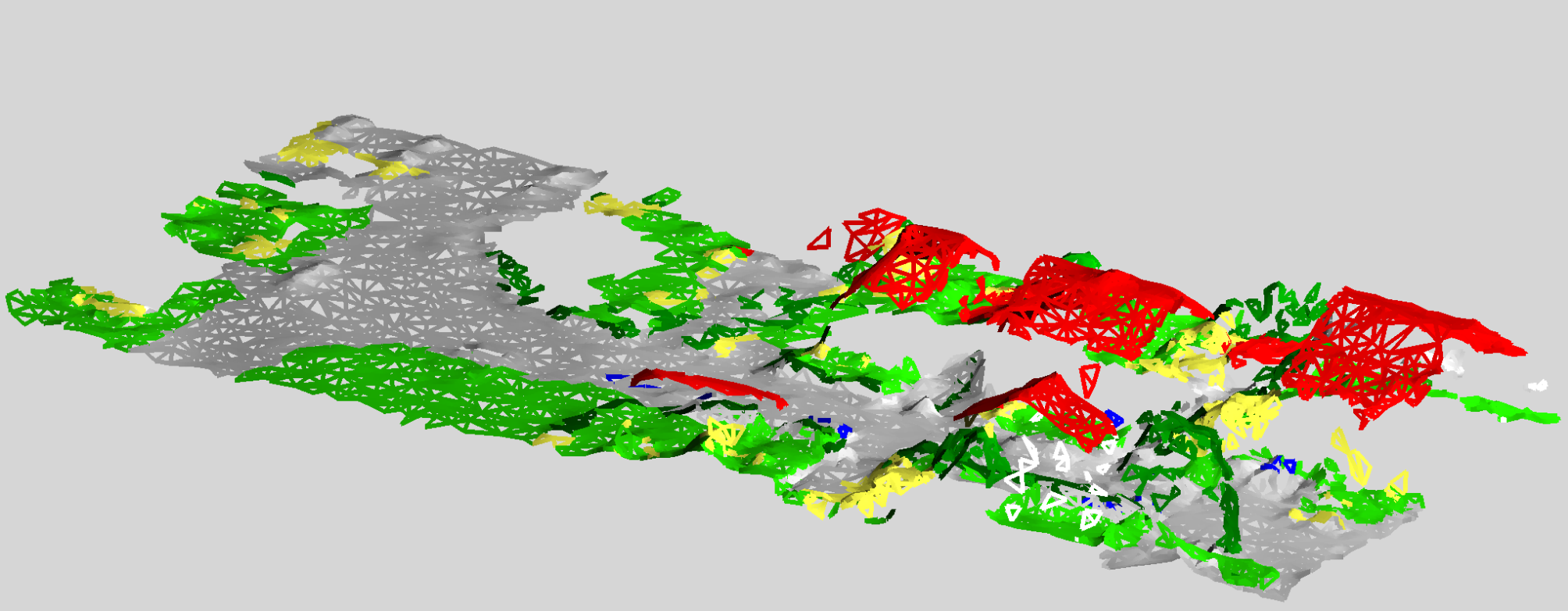}
    \end{tabular}
    \caption{Fringe area of \ac{V3D} represented by the \ac{PC} (\textit{top}) and the \ac{MVS} mesh (\textit{bottom}). The left column shows 3D data in height-coded fashion (\textit{blue}: low, \textit{red}: high). The top right shows the manually annotated \ac{PC}; the bottom right shows the automatically annotated mesh as wireframe. Faces that do not map the real geometry are not linked to the \ac{PC}, and hence, remain unlabeled (cf. holes in the leftmost part for tree and building).}  
    \label{fig:inconsistency_fringe}
\end{figure}

\autoref{fig:inconsistency_fringe} shows the fringe of the \ac{MVS} mesh (\textit{bottom}) and the respective \ac{ALS} \ac{PC} (\textit{top}) for \ac{V3D}. 
The height-coding indicates reconstruction errors in the leftmost part of the \ac{MVS} mesh:
the building and tree are not reconstructed. 
The adaptive thresholding guarantees a correct linking of faces and points where geometry is reconstructed correctly.  
Likewise, false reconstructed faces remain unlabeled and hence avoid label inconsistencies (after the backward pass). 

For \ac{V3D}, the adaptive thresholding associates 40.9\% of faces covering 53.8\% of the surface area with 75.6\% of LiDAR points. 
The proxy analysis reveals that 98.9\% of associated points show consistency in manual and back-transferred labels.
2.0\% of associated faces are linked to points of different classes causing label inconsistencies for 1.1\% of associated points. 
The achieved weighted average precision of the label consistency check is 98.9\%. 

For \ac{H3D}, the adaptive thresholding associates 67.3\% of faces covering 71.2\% of the surface area with 55.9\% of LiDAR points. 
99.6\% of points pass the label consistency check. 
Vice versa, 0.9\% of associated faces are linked to points of different classes. 
The achieved weighted average precision is 99.9\%. 

Structural differences among 3D modalities prevent full association of points and faces for both data sets (cf.~\autoref{fig:discrepancies_interrepresentation} and facades in~\autoref{fig:inconsistency_fringe}). 
Particularly semi-transparent objects reduce the association rates on the point-level since sub-surface LiDAR points are not linked to the mesh surface.
The majority of non-associated points belong to vegetational classes (\ac{V3D}: 68\%, \ac{H3D}: 74\%). 
In this regard, the high LiDAR density of \ac{H3D} causes a low association rate on the point-level. 
In contrast, the comparatively high association rate on the face-level indicates proper co-registration and high-quality mesh. 
The majority of non-associated faces build the water surface where no LiDAR points are captured. 
For \ac{V3D}, 15.5\% of non-linked points belong to \textit{facade} and \textit{roof} pointing out the minor \ac{MVS} mesh quality. 
To summarize, the association rates indicate the impact of the mesh quality and the point density.

\begin{figure}[htbp]
    \centering
    \includegraphics[width=0.8\columnwidth]{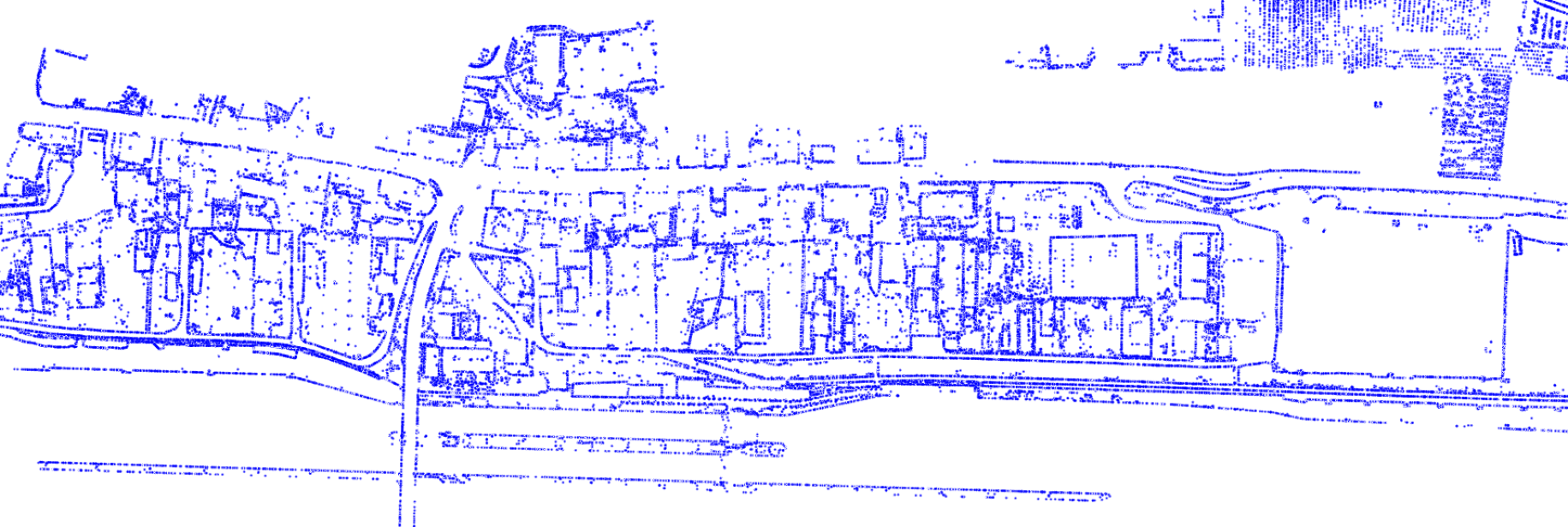}
    \vspace{0.1cm}
    \begin{tabular}{cc}
        \includegraphics[width=0.22\columnwidth]{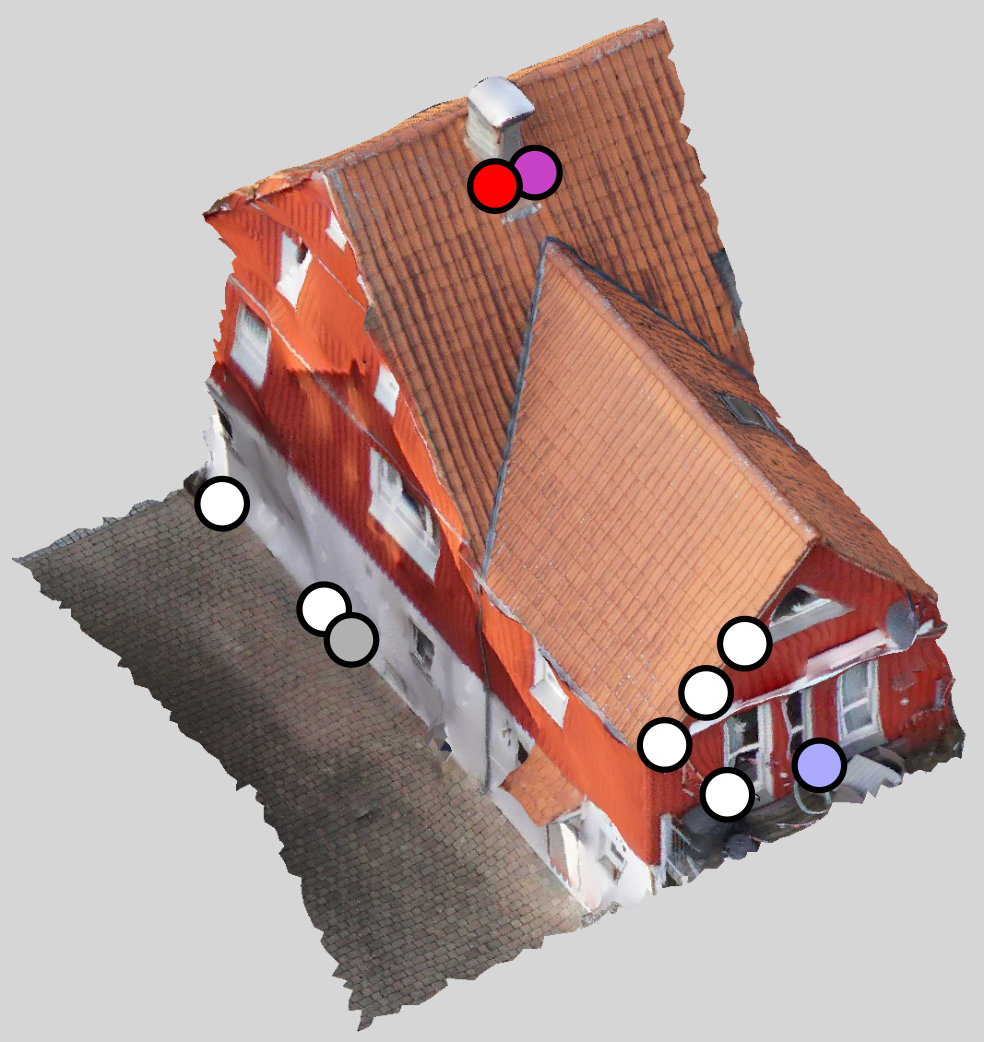} 
        \put (-56.25, -7.5) {\footnotesize \ac{GT} labels}
        &
        \includegraphics[width=0.22\columnwidth]{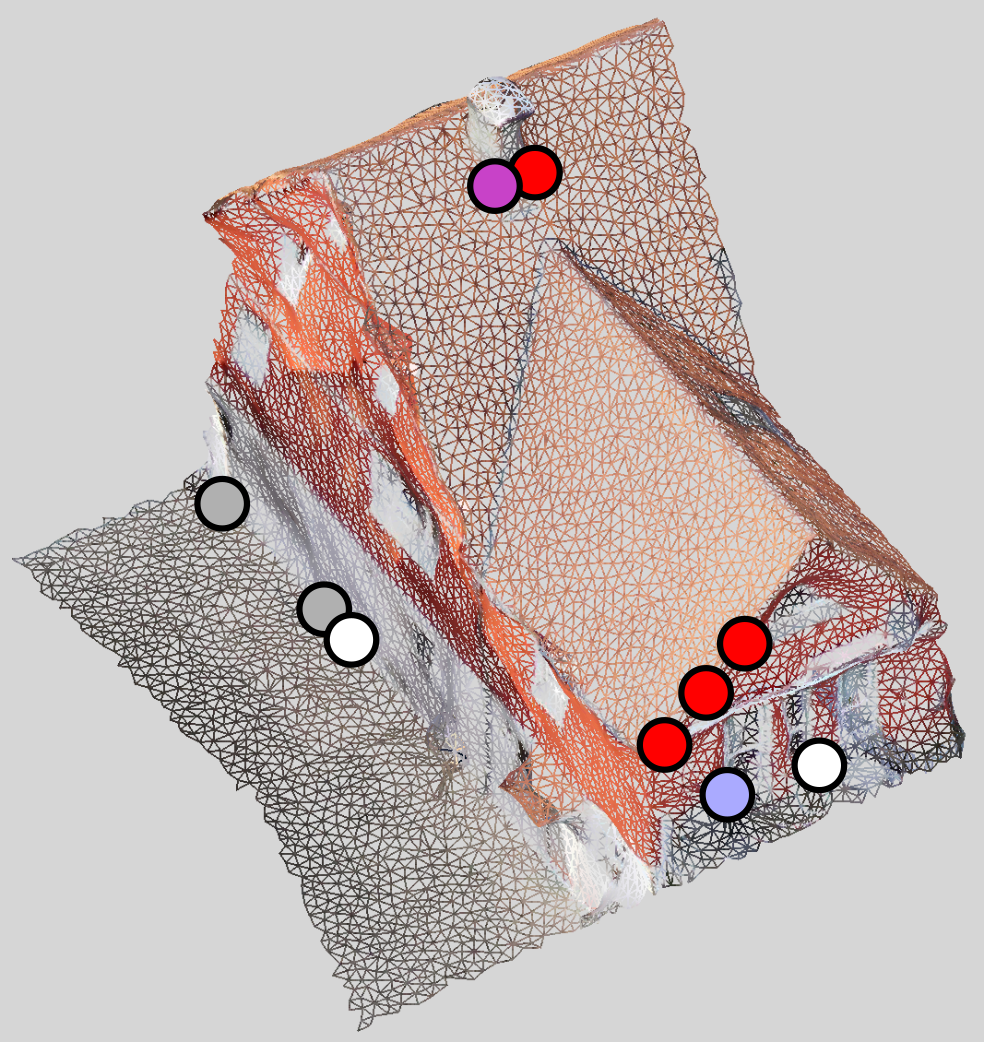} 
        \put (-80.25, -7.5) {\footnotesize back-transferred labels}
    \end{tabular}
    \caption{Overview of \ac{H3D} points that show inconsistencies in manual annotations and back-transferred labels (\textit{top}).
    The close-up at the bottom depicts inconsistently labeled points marked by the manually annotated \ac{GT} on the textured mesh (\textit{left}) and the back-transferred labels on the wireframe (\textit{right}).
    }
    \label{fig:inconsistency_overview}
\end{figure}

The proxy analysis reveals that the majority of established point-face connections is correct for both data sets (98.9\%/99.6\%).
Likewise, the forward-backward-pass shows that common meshing does not incorporate semantic borders. 
Hence, faces may be linked to points of different classes. In this case, the back-transferred majority vote causes inconsistencies.
\autoref{fig:inconsistency_overview} shows inconsistently labeled points for the entire \ac{H3D} data set and as close-up. 
The overview at the top exhibits 0.4\% of points failing the label consistency check.
These points represent semantic borders and indicate an improper mesh reconstruction. 
We are aware of the fact that co-registration discrepancies in 3D space increase this effect. 
However, due to the high mesh quality and small co-registration discrepancy, \ac{H3D} is less affected than \ac{V3D}.

\begin{figure}[htbp]
    \centering
    \includegraphics[width=0.85\columnwidth]{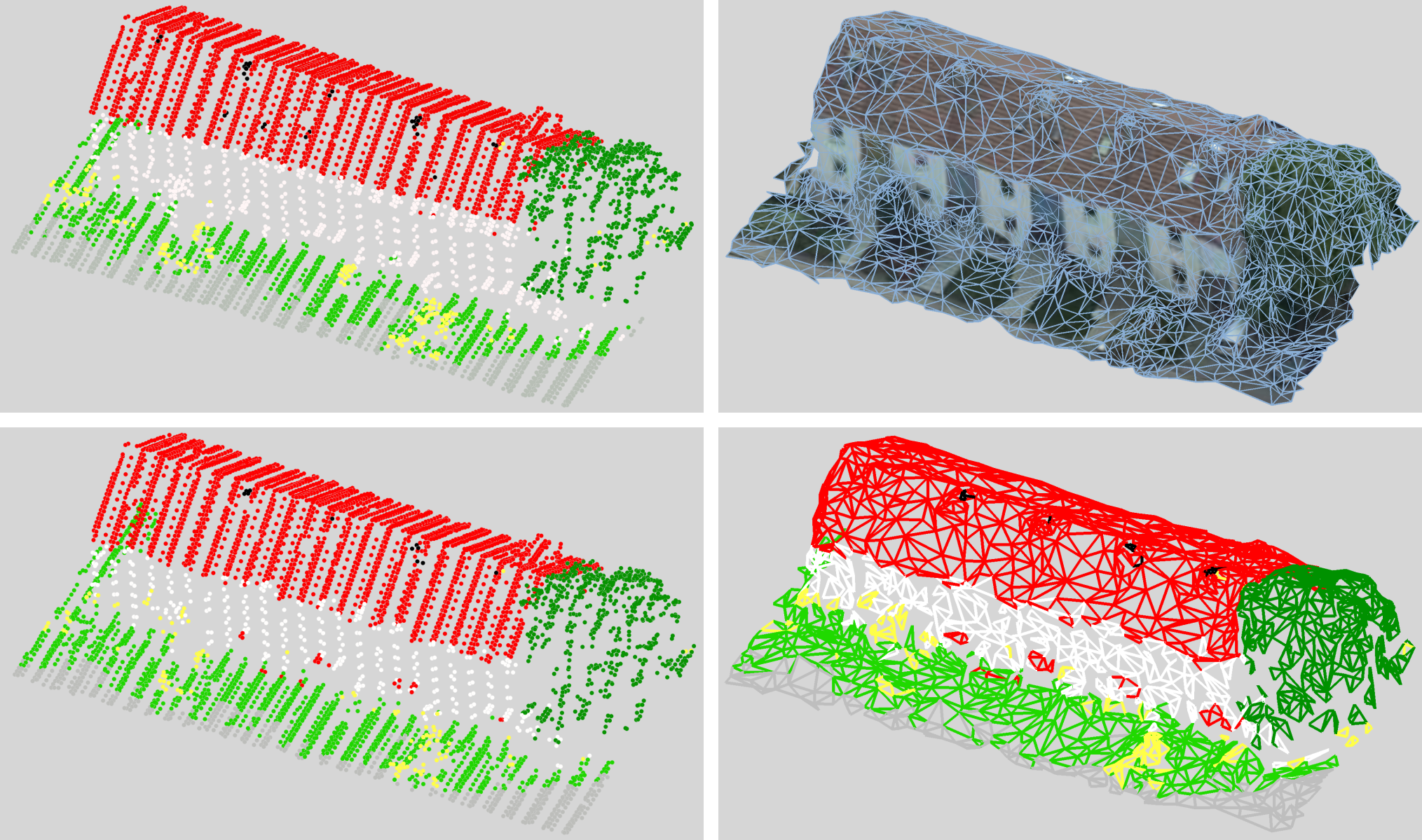}
    \caption{\ac{PC} (\textit{left}) and mesh representation (\textit{right}) of a building in \ac{V3D}. 
    The top row shows the manually annotated \ac{GT} and the textured mesh overlaid with its wireframe. 
    The bottom row shows the automatically labeled mesh (PC~$\mapsto$~Mesh) and the respectively labeled \ac{PC} with back-transferred labels from the mesh (Mesh~$\mapsto$~PC).}  
    \label{fig:inconsistency_building}
\end{figure}

\begin{figure}[htbp]
    \centering
    \includegraphics[width=0.55\columnwidth]{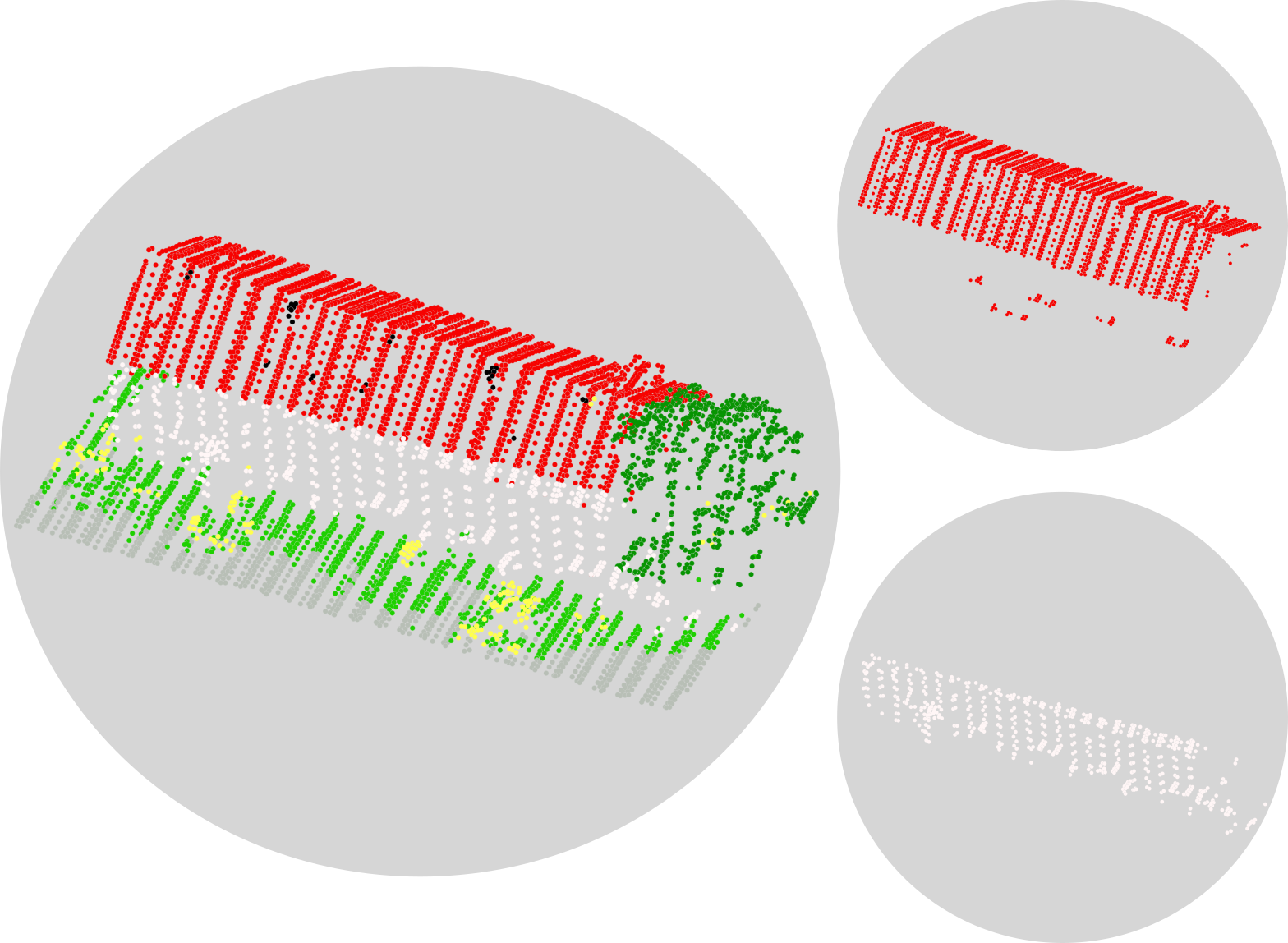}
    \caption{Label noise in form of duplicates in \ac{V3D} for building of \autoref{fig:inconsistency_building}. The right side shows \ac{GT} separated by class roof (\textit{top}) and facade (\textit{bottom})
    The label noise causes wrong label transfer and thus inconsistencies in the forward-backward-pass. 
    }
    \label{fig:inconsistency_building_label_noise}
\end{figure}

Furthermore, the proxy analysis helps to detect label noise. 
\autoref{fig:inconsistency_building} depicts a building from V3D as \ac{PC} (\textit{left}) and wireframe mesh (\textit{right}).
A few of the transferred labels to the mesh (\textit{lower right}) seem not to match the given \ac{GT} on the \ac{PC} (\textit{upper left}). 
For instance, some faces on the facade are marked as a roof.
Consequently, the back-transferred labels to the \ac{PC} (\textit{lower left}) do not match the initial annotation. 
Here, the appearance of false labels hints at label noise, since the inconsistencies cannot be explained by georeference issues. 
\autoref{fig:inconsistency_building_label_noise} shows the \ac{GT} along with its class-wise \ac{GT} for classes \textit{facade} and \textit{roof}. 
The figure discloses that few points erroneously carry labels of both classes.

\section{Conclusion and Future Work}
\label{sec:futurework}
To jointly leverage imagery and \ac{ALS} data for semantic scene analysis, we propose a novel holistic methodology that explicitly integrates imagery and \ac{PC} data via the mesh as the core representation.
The multi-modal data fusion establishes explicit connections of points, faces, and pixels and enables the subsequent sharing of arbitrary information across modalities. 
The information transfer incorporates the established one-to-many relationships by aggregation.
Hence, representation-specific features and (manual) annotations can be shared at a stroke across all modalities (cf.~\autoref{tab:association_options}). 
Therefore, the proposed association mechanism can be seen as an integrator that functions as a labeling tool and a feature sharing tool.
By these means, the novel method serves as a powerful integrative backbone boosting multi-modal learning.
In particular, the method underlines its utility for pixel-wise \ac{GT} generation. 
Any labeled 3D data (\ac{PC} or mesh) can be projected into image space to annotate multiple images at once while performing the visibility check. 
Hence, it minimizes the manual labeling effort. 
Consequently, the versatile applicability of the information transfer fosters modality-specific and multi-modal semantic segmentation.

The linking mechanism is designed to surpass imperfections of real-world data by adaptive thresholding.
The tile-wise parallel processing aims for a trade-off of memory and speed. 
We qualitatively and quantitatively demonstrate its effectiveness and adaptiveness to underlying data by deploying two airborne data sets of different resolutions and scales. 
\ac{V3D} is typical for large-scale country-wide mapping with moderate \ac{GSD} of some centimeters and a considerable time shift between \ac{ALS} and image data collection. 
\ac{H3D} provides extremely high-resolution data with mainly synchronous data capture from a hybrid sensor system and is representative of data collection at small-scale complex built-up areas. 
Due to the absence of multi-modal \ac{GT}, a strict quantitative analysis of the proposed method is difficult. 
As an alternative, we analyze the label consistency on the \ac{PC} by forward-backward-passes of labels across entities of different modalities. 
The quantitative analysis shows that nearly 100\% of the established connections are consistent. 
However, points of different classes that are linked to a common face might be useful for a subsequent semantically driven remeshing.
Due to structural discrepancies, full association across entities is not possible. 
We discuss preconditions and limitations in detail highlighting the benefits of high-quality co-registration and high-quality reconstruction. 
The strength of our method is its simplicity and flexibility that immediately profits from advances in data acquisition, co-registration, and meshing. 
In the future, we aim for pixel-accurate co-registration of \ac{ALS} data and imagery leveraging the hybrid strip adjustment \citep{glira2019}. 
We claim that improved co-registration improves both geometric reconstruction and semantic analysis. 
To prove our assumption, we plan an ablation study for multi-modal features on different representations by analyzing the performance of a \ac{ML} classifier.

\section{Acknowledgements}
\ac{V3D} is provided by the German Society for Photogrammetry, Remote Sensing and Geoinformation (DGPF) \citep{Cramer2010}. 
\ac{H3D} data originates from a research project in collaboration with the German Federal Institute of Hydrology~(BfG) in~Koblenz.
We thank all our colleagues 
for insightful discussions.
In particular, we thank our students Fangwen Shu, Mohamad Hakam Shams Eddin, and Vishal Pani, who assisted the implementation.
Furthermore, we thank the whole nFrames team for their support regarding the mesh generation.
Special thanks are directed to Carmen Kaspar for proofreading.

\bibliography{mybibfile}
\let\thefootnote\relax\footnotetext{\hspace{-0.55cm}\textbf{Copyright:} © 2021. This manuscript version is made available under the CC-BY-NC-ND 4.0 license \url{http://creativecommons.org/licenses/by-nc-nd/4.0/}.}

\end{document}